\documentclass{article}

\PassOptionsToPackage{sort}{natbib}

\usepackage[final]{neurips_2019}

\usepackage[utf8]{inputenc} 
\usepackage[T1]{fontenc}    
\usepackage[]{hyperref}       
\hypersetup{
colorlinks=true,
linkcolor=blue,         
citecolor=[rgb]{0,0,0.75},
}

\usepackage{url}            
\usepackage{booktabs}       
\usepackage{amsfonts}       
\usepackage{nicefrac}       
\usepackage{microtype}      

\usepackage{dsfont}
\usepackage{definitions}

\usepackage{xcolor}
\usepackage{amsfonts}     
\usepackage{amsmath}
\usepackage{amsthm}
\usepackage{amssymb}

\usepackage{multirow}
\usepackage{graphicx}
\usepackage{rotating}

\usepackage{tikz}
\usetikzlibrary{arrows}
\definecolor{out-col}{RGB}{230,230,230}
\definecolor{H1-1-col}{RGB}{252,141,98}
\definecolor{H1-2-col}{RGB}{102,194,165}
\definecolor{H1-3-col}{RGB}{141,160,203}
\definecolor{H2-1-col}{RGB}{255,217,47}
\definecolor{H2-2-col}{RGB}{166,216,84}
\definecolor{H2-3-col}{RGB}{231,138,195}

\usetikzlibrary{positioning,chains,arrows}
\usetikzlibrary{decorations.pathmorphing}
\usetikzlibrary{decorations.markings}
\usetikzlibrary{decorations.pathreplacing}
\usetikzlibrary{calc}
\usetikzlibrary{arrows.meta}
\usetikzlibrary{bending}
\usetikzlibrary{intersections}
\usetikzlibrary{fadings}
\usetikzlibrary{babel}
\usetikzlibrary{shapes}

\tikzstyle{tips}=[%
btip/.tip = {Latex[length=6, width=3.5]},
-btip]
\tikzstyle{neuralnetwork}=[%
draw=black,
node distance = 3mm and 8mm,
start chain = going right,
neuron/.style = {circle, draw=black,
	minimum size=13pt, inner sep=0pt,
	on chain},
annot/.style = {text width=4em, align=center,},
tips]

\usepackage{afterpage}
\usepackage[textsize=scriptsize,textwidth=2cm]{todonotes}

\newtheorem{thm}{Theorem}

\newtheorem{lem}[thm]{Lemma}

\theoremstyle{definition}

\theoremstyle{plain}

\usepackage{bbold}

\usepackage[nameinlink]{cleveref}

\title{Dichotomize and Generalize: PAC-Bayesian Binary Activated Deep Neural Networks}

\author{%
  Ga\"el Letarte\\
  Universit\'e Laval\\
  Canada\\
  \texttt{gael.letarte.1@ulaval.ca}\\
  \And
  Pascal Germain\\
  Inria \\
  France\\
  \texttt{pascal.germain@inria.fr}\\
  \AND
  Benjamin Guedj\\
  Inria and University College London \\
  France and United Kingdom \\
  \texttt{benjamin.guedj@inria.fr} \\
  \And
  Fran\c cois Laviolette \\
  Universit\'e Laval\\
  Canada\\
  \texttt{francois.laviolette@ift.ulaval.ca} \\
}

\begin{document}

\maketitle

\begin{abstract}
We present a comprehensive study of multilayer neural networks with binary activation, relying on the PAC-Bayesian theory.
Our contributions are twofold: 
(i) we develop an end-to-end framework to train a binary activated deep neural network, 
(ii) we provide nonvacuous PAC-Bayesian generalization bounds for binary activated deep neural networks. 
Our results are obtained by minimizing the expected loss of an architecture-dependent aggregation of binary activated deep neural networks.
Our analysis inherently overcomes the fact that binary activation function is non-differentiable.
The performance of our approach is assessed on a thorough numerical experiment protocol on real-life datasets.

\end{abstract}


\section{Introduction}\label{sec:intro}

The remarkable practical successes of deep learning make the need for better theoretical understanding all the more pressing.
The PAC-Bayesian theory has recently emerged as a fruitful framework to analyze generalization abilities of deep neural network. Inspired by precursor work of \citet{langford2001not}, nonvacuous risk bounds for multilayer architectures have been obtained by \citet{Karolina2017,zhou2018nonvacuous}. 
Although informative, these results do not explicitly take into account the network architecture (number of layers, neurons per layer, type of activation function). 
A notable exception is the work of \citet{neyshabur18} which provides a PAC-Bayesian analysis relying on the network architecture and the choice of ReLU activation function. The latter bound arguably gives insights on the generalization mechanism of neural networks (namely in terms of the spectral norms of the learned weight matrices), but their validity hold for some margin assumptions, and they are likely to be numerically vacuous.

We focus our study on deep neural networks with a sign activation function. We call such networks \emph{binary activated multilayer} (BAM) networks. This specialization leads to nonvacuous generalization bounds which hold under the sole assumption that training samples are \iid. We provide a PAC-Bayesian bound holding on the generalization error of a continuous aggregation of BAM networks. This leads to an original approach to train BAM networks, named PBGNet.
The building block of PBGNet arises from the specialization of PAC-Bayesian bounds to linear classifiers \citep{germain2009pac}, that we adapt to deep neural networks. 
The term \emph{binary neural networks} has been coined by \citet{bengio09}, and further studied in \cite{hubara2016binarized,DBLP:conf/nips/SoudryHM14,hubara2017quantized}: it refers to neural networks for which both the activation functions and the weights are binarized (in contrast with BAM networks). These architectures are motivated by the desire to reduce the computation and memory footprints of neural networks. 

Our theory-driven approach is validated on real life datasets, showing competitive accuracy with $\tanh$-activated multilayer networks, and providing nonvacuous generalization bounds.




\textbf{Organisation of the paper.}
We formalize our framework and notation in Section \ref{sec:notation}, along with a presentation of the PAC-Bayes framework and its specialization to linear classifiers.
Section~\ref{sec:twolayers} illustrates the key ideas we develop in the present paper, on the simple case of a two-layers neural network. This is then generalized to deep neural networks in Section~\ref{sec:multilayers}. We present our main theoretical result in Section~\ref{sec:pbgnet}: a PAC-Bayesian generalization bound for binary activated deep neural networks, and the associated learning algorithm. Section~\ref{sec:xp} presents the numerical experiment protocol and results. The paper closes with avenues for future work in Section~\ref{sec:conclusion}.

\section{Framework and notation}\label{sec:notation}

We stand in the supervised binary classification setting: given a real input vector\footnote{Bold uppercase letters denote matrices, bold lowercase letters denote vectors.} $\xbf\in\Rds^{d_0}$, one wants to predict a label $y\in\{-1,1\}$.
Let us consider a neural network of $L$ \emph{fully connected} layers with a (binary) sign activation function: $\sgn(a) = 1$ if $a > 0$ and $\sgn(a)=-1$ otherwise.\footnote{We consider the activation function as an \emph{element-wise} operator when applied to vectors or matrices.}
We let $d_k$ denote the number of neurons of the $k^{\rm th}$ layer, for $k \in\{1,\ldots,L\}$; $d_0$ is the input data point dimension, and $D\eqdots \sum_{k=1}^L d_{k-1}d_k$ is the total number of parameters.
The output of the (deterministic) BAM network on an input data point $\xbf\in\Rds^{d_0}$ is given by
\begin{align}\label{eq:BAM}
    f_\theta(\xbf) = \sgn\big(\Wk[L] \sgn\big(\Wk[L-1] \sgn\big(\ldots \sgn\big(\Wk[1]\xbf\big)\big)\big)\big)\,,
\end{align}
where $\Wk\in \Rds^{d_{k}\times d_{k-1}}$ denotes the weight matrices.
%
%
The network is thus parametrized by $\theta \,{=}\, \vec\!
\left(\{ \Wk\}_{k=1}^{L}\right){\in}\Rds^D$. The $i^{\rm th}$ line of matrix $\Wbf_k$ will be denoted $\wbf_k^i$.
For binary classification, the BAM network final layer $\Wk[L]{\in}\Rds^{1\times d_{L-1}}$ has one line ($d_L{=}1$), that is a vector $\wbf_L{\in}\Rds^{d_{L-1}}$, and 
$f_\theta:\Rds^{d_0}{\to}\{-1,1\}$.


\subsection{Elements from the PAC-Bayesian theory}
The Probably Approximately Correct (PAC) framework \citep[introduced by][]{valiant1984theory} holds under the frequentist assumption that data is sampled in an \iid{} fashion from a data distribution $\Dcal$ over the input-output space.
The learning algorithm observes a finite  training sample $S=\{(\xbf_i,y_i)\}_{i=1}^n\sim \Dcal^{n}$ and outputs a predictor $f:\Rds^{d_0}\to [-1,1]$.
Given a loss function $\ell:[-1,1]^2\to [0,1]$,
we define $\Lcal_\Dcal(f) $ as the  generalization loss on the data generating distribution $\Dcal$, and $\widehat\Lcal_S(f) $ as the empirical error on the training set, given by
\begin{align*}
\Lcal_\Dcal (f) =  \Esp_{(\xbf,y)\sim \Dcal} \ell(f(\xbf), y)\,,
\quad \mbox{ and }\quad
\widehat\Lcal_S (f) = \frac1n \sum_{i=1}^n \ell(f(\xbf_i), y_i)\,.
\end{align*}
PAC-Bayes considers the expected loss of an aggregation of predictors: considering a distribution~$Q$ (called the \emph{posterior}) over a family of predictors $\Fcal$, one obtains PAC upper bounds on 
$\Esp_{f\sim Q} \Lcal_\Dcal(f)$. Our work focuses on the linear loss
$\ell(y',y) \eqdots \frac12(1-y y')$, for which the aggregated loss is equivalent to the loss of the predictor 
$F_Q(\xbf) \eqdots \Esp_{f\sim Q} f(\xbf)$,
performing a $Q$-aggregation of all predictors in~$\Fcal$. 
In other words, we may upper bound with an arbitrarily high probability the generalization loss $\Lcal_\Dcal(F_Q) = \Esp_{f\sim Q} \Lcal_\Dcal(f)$, by its empirical counterpart 
$\widehat\Lcal_S(F_Q)= \Esp_{f\sim Q} \widehat\Lcal_S(f)$
and a complexity term, the Kullback-Leibler divergence between $Q$ and a reference measure~$P$ (called the \emph{prior} distribution)  chosen independently of the training set~$S$, given by
$\KL(Q\|P) \eqdots \int  \ln \frac{Q(\theta)}{P(\theta)}Q(\mathrm{d}\theta) $.
Since the seminal works of \citet{shawetaylor-97}, \citet{mcallester-99,mcallester-03a} and \citet{catoni2004statistical,catoni2003pac,catoni-07}, the celebrated PAC-Bayesian theorem has been declined in many forms \citep[see][for a survey]{guedj2019primer}. The following Theorems~\ref{thm:pbseeger} and~\ref{thm:catoni} will be useful in the sequel.
\pagebreak

\begin{thm}[\citet{seeger-02}, \citet{maurer-04}] \label{thm:pbseeger}
Given a prior $P$ on $\Fcal$, with probability at least $1-\delta$ over $S\sim \Dcal^{n}$, 
\begin{align} \label{eq:seeger}
   \mbox{for all $Q$ on $\Fcal$\,:}\quad
   \kl\Big(\widehat\Lcal_S (F_Q)\big\|\Lcal_\Dcal (F_Q)\Big) \ \leq \ 
  \frac{\KL(Q\|P)+\ln\frac{2\sqrt{n}}{\delta}}{n}\,,
\end{align}
where $\kl(q\|p) \eqdots q\ln\frac{q}{p} + (1-q)\ln\frac{1-q}{1-p}$ is the Kullback-Leibler divergence between Bernoulli distributions with probability of success $p$ and $q$, respectively.
\end{thm}
\begin{thm}[\citet{catoni-07}]  \label{thm:catoni}
Given $P$ on $\Fcal$ and $C>0$, with probability at least $1-\delta$ over $S\sim \Dcal^{n}$, 
\begin{align} \label{eq:catoni}
   \mbox{for all $Q$ on $\Fcal$\,:}\quad
   \Lcal_\Dcal (F_Q) \leq \ 
\frac{1}{1-e^{-C}}\left(
     1-\exp\left(-C \, \widehat\Lcal_S (F_Q) - \frac{\KL(Q\|P)+\ln\frac{1}{\delta}}{n}\right)
     \right)   .
\end{align}
\end{thm}
From Theorems~\ref{thm:pbseeger} and~\ref{thm:catoni}, we obtain PAC-Bayesian bounds on the \emph{linear loss} of the $Q$-aggregated predictor $F_Q$. 
Given our binary classification setting, it is natural to predict a label by taking the sign of $F_Q(\cdot)$. Thus, one may also be interested in the \emph{zero-one loss} $ \ell_{\rm 01} (y', y) \eqdef \mathbb{1}[\sgn(y')\neq y]$; the bounds obtained from Theorems~\ref{thm:pbseeger} and~\ref{thm:catoni} can be turned into bounds on the \emph{zero-one loss} with an extra~$2$ multiplicative factor, using the elementary inequality 
$ \ell_{\rm 01} (F_Q(\xbf), y) \leq 2\ell(F_Q(\xbf),y)$.

\subsection{Elementary building block: PAC-Bayesian learning of linear classifiers}
\label{section:pbgd}

The PAC-Bayesian specialization to linear classifiers has been proposed by~\citet{langford-02}, and used for providing tight generalization bounds and a model selection criteria \citep[further studied by][]{langford-05,ambroladze-06,parrado-12}. This paved the way to the PAC-Bayesian bound minimization algorithm of \citet{germain2009pac},  that learns a linear classifier $f_\wbf(\xbf) \eqdef \sgn(\wbf\cdot\xbf)$, with $\wbf\in\Rds^{d}$. The strategy is to consider a Gaussian posterior $Q_\wbf \eqdef \Ncal(\wbf, I_{d})$ and a Gaussian prior $P_{\wbf_0}\eqdef \Ncal(\wbf_0, I_{d})$ over the space of all linear predictors $\Fcal_d \eqdef \{ f_\vbf | \vbf \in \Rds^d\}$ (where $I_d$ denotes the $d\times d$ identity matrix).
The posterior is used to define a linear predictor $f_\wbf$ and the prior may have been learned on previously seen data; a common uninformative prior being the null vector $\wbf_0 = \zerobf$. With such parametrization, $\KL(Q_\wbf \| P_{\wbf_\zerobf})= \frac12 \|\wbf- \wbf_0\|^2$. Moreover, the  $Q_\wbf$-aggregated output can be written in terms of the Gauss error function $\Erf(\cdot)$.
In \citet{germain2009pac}, the erf function is introduced as a loss function to be optimized. Here we interpret it as the predictor output, to be in phase with our neural network approach. Likewise, we study the linear loss of an aggregated predictor instead of the \emph{Gibbs risk} of a stochastic classifier. We obtain (explicit calculations are provided in Appendix~\ref{sec:sgn2erf} for completeness)
\begin{equation} \label{eq:erfactivated}
F_\wbf(\xbf)
\eqdef
\Esp_{\vbf\sim Q_\wbf} f_\vbf(\xbf)
\, =  \,
\Erf\!\left( \tfrac{\wbf\cdot\xbf}{\sqrt{2} \|\xbf\|} \right),
\quad \mbox{with }\ \Erf(x) \eqdef \tfrac{2}{\sqrt{\pi}}\textstyle\int_{0}^{x}e^{-t^{2}}dt\,.    
\end{equation}
Given a training set $S\sim \Dcal^{n}$, \citet{germain2009pac} propose to minimize a PAC-Bayes upper bound on $\Lcal_\Dcal(F_\wbf)$ by gradient descent on $\wbf$. This approach is appealing as the bounds are valid uniformly for all $Q_\wbf$ (see Equations~\ref{eq:seeger} and~\ref{eq:catoni}). In other words, the algorithm provides both a learned predictor and a generalization guarantee that is rigorously valid (under the \iid{} assumption) even when the optimization procedure did not find the global minimum of the cost function (either because it converges to a local minimum, or early stopping is used). \citet{germain2009pac} investigate the optimization of several versions of 
Theorems~\ref{thm:pbseeger} and~\ref{thm:catoni}. The minimization of Theorem~\ref{thm:pbseeger} generally leads to tighter bound values, but empirical studies show lowest accuracy as the procedure conservatively prevents overfitting. The best empirical results are obtained by minimizing Theorem~\ref{thm:catoni} for a fixed hyperparameter $C$, selected by cross-validation. Minimizing Equation~\eqref{eq:catoni} amounts to minimizing
\begin{align} \label{eq:pbgd3}
     C \,n\, \widehat\Lcal_S (F_\wbf) + \KL(Q_\wbf\|P_{\wbf_0})
     \ = \
      C \, \frac 12\sum_{i=1}^n \Erf\!\left(-y_i\, \frac{\wbf\cdot\xbf_i}{\sqrt{2} \|\xbf_i\|} \right) + \frac 12\|\wbf- \wbf_0\|^2\,.
\end{align}
In their discussion, \citet{germain2009pac} observe that the objective in Equation~\eqref{eq:pbgd3} is similar to the one optimized by the soft-margin Support Vector Machines \citep{cortes-95}, by roughly interpreting the \emph{hinge loss} $\max(0, 1-yy')$ as a convex surrogate of the \emph{probit loss }$\Erf(-yy')$. Likewise, \citet{langford-02} present this parameterization of the PAC-Bayes theorem as a margin bound. In the following, we develop an original approach to neural networks based on a slightly different observation: the predictor output given by Equation~\eqref{eq:erfactivated} is reminiscent of the $\tanh$ activation used in classical neural networks (see Figure~\ref{fig:act_func} in the appendix for a visual comparison). 
Therefore, as the linear \emph{perceptron} is viewed as the \emph{building block} of modern multilayer neural networks, the PAC-Bayesian specialization to binary classifiers is the cornerstone of our theoretical and algorithmic framework for BAM networks.

\section{The simple case of a one hidden layer network}\label{sec:twolayers}
Let us first consider a network with one hidden layer of size $d_1$. Hence, this network is parameterized by weights $\theta=\vec(\{\Wbf_1, \wbf_2\})$, with $\Wbf_1\in\Rds^{d_1\times d_0}$ and $\wbf_2\in\Rds^{d_1}$. Given an input $\xbf\in\Rds^{d_0}$, the output of the network is
\begin{equation} \label{eq:2BAM}
    f_{\theta}(\xbf) = \sgn\big(\wbf_2 \cdot \sgn(\Wbf_1 \xbf)\big)\,.
\end{equation}
Following Section~\ref{sec:notation}, we consider an isotropic Gaussian posterior distribution centered in $\theta$, denoted $Q_\theta = \Ncal(\theta, I_D)$, over the family of all networks $\Fcal_D = \{ f_{\tilde\theta}\, |\, \tilde\theta \in \Rds^D\}$. Thus, the prediction of the $Q_\theta$-aggregate predictor is given by 
$ F_\theta(\xbf) =  \Esp_{\tilde\theta \sim Q_\theta} f_{\tilde\theta} ( \xbf) $. 
Note that \citet{langford2001not,Karolina2017} also consider Gaussian distributions over neural networks parameters. However, as their analysis is not specific to a particular activation function---experiments are performed with \emph{typical} activation functions (sigmoid, ReLU)---the prediction relies on sampling the parameters according to the posterior. An originality of our approach is that, by studying the sign activation function, we can calculate the exact form of $F_\theta(\xbf)$, as detailed below.

\subsection{Deterministic network}
\label{sec:derterministic_network}

\textbf{Prediction.}
To compute the value of $F_\theta(\xbf)$, we first need to decompose the probability of each $\tilde\theta {=} \vec(\{\Vbf_1, \vbf_2\}) \,{\sim}\, Q_\theta$ as
$Q_\theta(\tilde\theta) {=} Q_{1}(\Vbf_1) Q_{2}(\vbf_2)$, with $Q_{1}{=} \Ncal(\Wbf_1, I_{d_0 d_1})$ and  $Q_{2}{=}\Ncal(\wbf_2, I_{d_1})$.
%
\begin{align}
    F_\theta(\xbf) 
    = &
    \int_{\Rds^{d_1\times d_0}} Q_1 (\Vbf_1) \int_{\Rds^{d_1}} Q_2 (\vbf_2) \sgn(\vbf_2 \cdot \sgn( \Vbf_1 \xbf)) d\vbf_2 d\Vbf_1
    \nonumber \\
    =&  \int_{\Rds^{d_1\times d_0}} Q_1 (\Vbf_1) \, \Erf \left( \tfrac{\wbf_2 \cdot \sgn(\Vbf_1 \xbf)}{\sqrt{2} \| \sgn( \Vbf_1\xbf)\|} \right) d \Vbf_1
    \label{eq:Q2erf}\\
    =& \sum_{\sbf\in\{-1,1\}^{d_1}} \Erf \left( \tfrac{\wbf_2 \cdot \sbf}{\sqrt{2d_1}} \right) \int_{\mathrlap{\Rds^{_{d_1\times d_0}}}} \  \mathbb{1}[\sbf=\sgn( \Vbf_1\xbf)] Q_1(\Vbf_1) \,  d \Vbf_1
    \label{eq:combine}
    \\
    =& \sum_{\sbf\in\{-1,1\}^{d_1}} \Erf \left( \tfrac{\wbf_2 \cdot \sbf}{\sqrt{2d_1}} \right) 
    \Psi_\sbf\left(\xbf, \Wbf_1 \right),
    \label{eq:F2layers}
\end{align}
where, from $Q_1(\Vbf_1) = \prod_{i=1}^{d_1} Q_1^i(\vbf_1^i)$ with $Q_1^i\eqdots\Ncal(\wbf_1^i, I_{d_0})$, we obtain
\begin{align} \label{eq:psi1}
\Psi_\sbf\left(\xbf, \Wbf_1 \right) \ \eqdef \ 
 \prod_{i = 1}^{d_1} \int_{\mathrlap{\Rds^{_{d_0}}}} \ \mathbb{1}[s_i\, \xbf\cdot\vbf_1^i > 0] Q_1^i(\vbf_1^i) \,  d \vbf_1^i
 \ = \ 
\prod_{i = 1}^{d_1} 
\underbrace{
\left[\frac{1}{2} + 
\frac{s_i}{2}\,\Erf\!\left(\frac{\wbf_1^i \cdot \xbf}{\sqrt{2}\norm{\xbf}}\right)  \right]}_{\psi_{s_i} (\xbf, \wbf^i_1)} .
\end{align}
Line~\eqref{eq:Q2erf} states that the output neuron is a linear predictor over the hidden layer's activation values $\sbf=\sgn(\Vbf_1 \xbf)$; based on Equation~\eqref{eq:erfactivated}, the integral on $\vbf_2$ becomes $\Erf \big( \wbf_2 \cdot     \sbf / (\sqrt{2}\| \sbf\|) \big)$.
As a function of $\sbf$, the latter expression is piecewise constant. Thus, Line~\eqref{eq:combine} discretizes the integral on~$\Vbf_1$ as a sum of the $2^{d_1}$ different values of $\sbf=(s_i)_{i=1}^{d_1}, s_i\in\{-1,1\}$. Note that $\|\sbf\|^2=d_1$.

Finally, one can compute the exact output of
$F_\theta(\xbf)$, provided one accepts to compute a sum combinatorial in the number of hidden neurons (Equation~\ref{eq:F2layers}).
We show in forthcoming Section \ref{section:stochastic} that it is possible to circumvent this computational burden and approximate $F_\theta(\xbf)$ by a sampling procedure.

\textbf{Derivatives.}
Following contemporary approaches in deep neural networks \citep{Goodfellow-16-book}, we minimize the empirical loss  $\widehat\Lcal_S(F_\theta)$ by stochastic gradient descent (SGD). This requires to compute the partial derivative of the cost function according to the parameters $\theta$:
\begin{align}
 \frac{\partial \widehat\Lcal_S(F_\theta)}{\partial \theta}
 = 
\frac1n \sum_{i=1}^n \frac{\partial \ell(F_\theta(\xbf_i), y_i)}{\partial \theta}
= 
\frac1n \sum_{i=1}^n 
\frac{\partial F_\theta(\xbf_i)}{\partial \theta}\,
\ell'(F_\theta(\xbf_i), y_i)\,,
\end{align}
with the derivative of the linear loss  $\ell'(F_\theta(\xbf_i), y_i)= -\frac12 y$.

The partial derivatives of the prediction function (Equation~\ref{eq:F2layers}) according to the hidden layer parameters $\wbf_1^k \in \{\wbf_1^1, \ldots, \wbf_1^{d_1}\}$ and the output neuron parameters $\wbf_2$ are
\begin{align} 
\frac{\partial }{\partial\wbf_1^k} F_\theta(\xbf)
    =& 
    \frac{\xbf}{2^{\frac32}\|\xbf\|}
    \Erf'\left(\frac{\wbf_1^k \cdot \xbf}{\sqrt{2}\norm{\xbf}}\right)  \sum_{\sbf\in\{-1,1\}^{d_1}} s_k\, \Erf \left( \frac{\wbf_2 \cdot \sbf}{\sqrt{2d_1}} \right) \left[\frac{\Psi_\sbf (\xbf, \Wbf_1)}{\psi_{s_k} (\xbf, \wbf_1^k)}
    \right],
    \label{eq:F2dw1}
    \\
\frac{\partial }{\partial\wbf_2} F_\theta(\xbf)
=& \frac{1}{\sqrt{2d_1}}  \sum_{\sbf\in\{-1,1\}^{d_1}}
\sbf \, \Erf' \left( \frac{\wbf_2 \cdot \sbf}{\sqrt{2d_1}} \right) 
   \Psi_\sbf(\xbf, \Wbf_1 )\,, 
   \quad
   \mbox{with \ $\Erf'(x)\eqdots \frac{2}{\sqrt \pi} e^{-x^2}$\,.}
   \label{eq:F2dw2}
\end{align}
Note that this is an exact computation. A  salient fact is that even though we work on non-differentiable BAM networks, we get a 
structure trainable by (stochastic) gradient descent by aggregating 
networks. 

\begin{figure}[t]
  \centering
  \includegraphics[trim={0.2cm 0.1cm 0.1cm 0cm},clip,width=\linewidth]{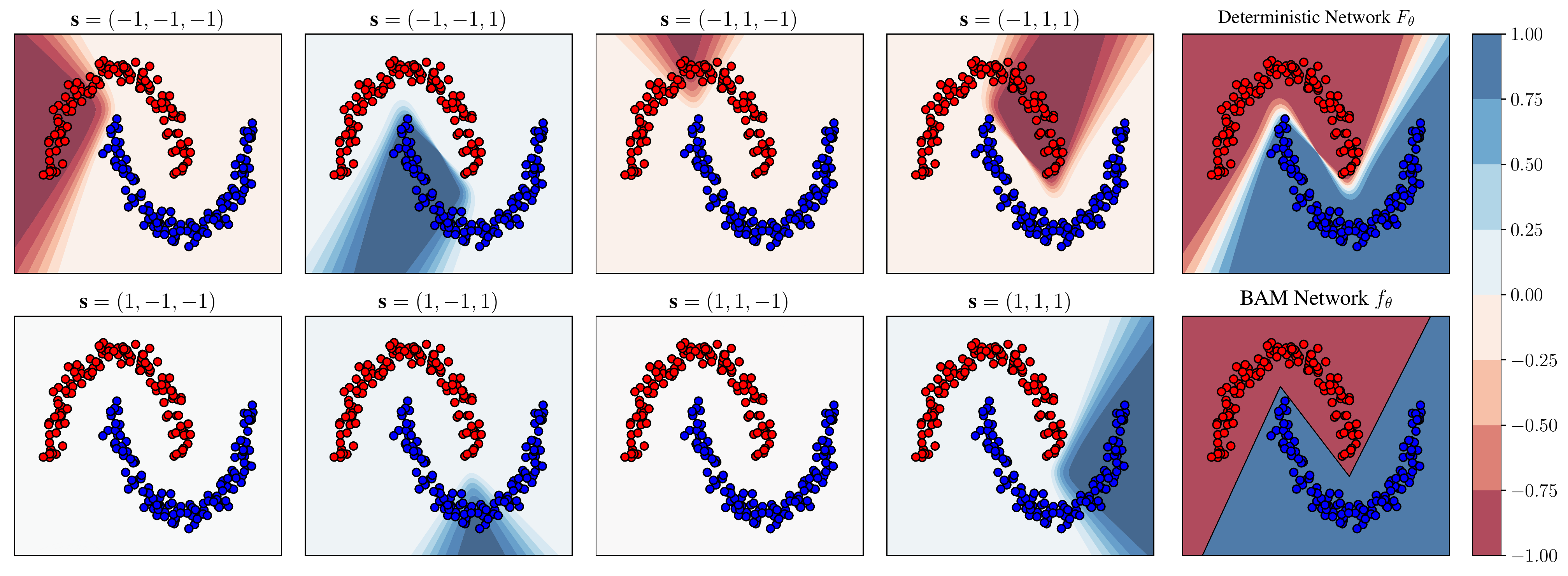}
  \caption{Illustration of the proposed method for a one hidden layer network of size $d_1{=}3$, interpreted as a majority vote  over $8$ binary representations $\sbf\in\{-1,1\}^{3}$. For each $\sbf$, a plot shows the values of $F_{\wbf_2}(\sbf) \Psi_\sbf(\xbf,\Wbf_1)$. The sum of these values gives the deterministic network output $F_\theta(\xbf)$ (see Eq.~\ref{eq:F2layers}). We also plot the BAM network output $f_\theta(\xbf)$ for the same parameters $\theta$ (see Eq.~\ref{eq:2BAM}).}
   \label{fig:moons}
\end{figure}

\textbf{Majority vote of learned representations.}
Note that $\Psi_\sbf$ (Equation~\ref{eq:psi1}) defines a distribution on~$\sbf$. Indeed, $\sum_\sbf \Psi_\sbf(\xbf, \Wbf_1){=}1$, as 
$\Psi_\sbf(\xbf, \Wbf_1 ) + \Psi_{\bar\sbf}(\xbf, \Wbf_1) = 2^{-d_1}$ for every $\bar\sbf\,{=}\,{-}\sbf$.
Thus, by Equation~\eqref{eq:F2layers} we can interpret $F_\theta$ akin to a majority vote predictor, which performs a convex combination of a linear predictor outputs $F_{\wbf_2}(\sbf)\eqdef\Erf ( \wbf_2 \cdot \sbf / \sqrt{2d_1} )$. The vote aggregates the predictions on the $2^{d_1}$ possible binary representations. 
Thus, the algorithm does not learn the representations \emph{per se}, but rather the weights $\Psi_\sbf(\xbf,\Wbf_1)$ associated to every $\sbf$ given an input $\xbf$, as illustrated by Figure \ref{fig:moons}.

\subsection{Stochastic approximation}
\label{section:stochastic}

Since $\Psi_\sbf$ (Equation~\ref{eq:psi1}) defines a distribution, we can interpret the function value as the probability of mapping input~$\xbf$ into the hidden representation $\sbf$ given the parameters $\Wbf_1$. Using a different formalism, we could write $\Pr(\sbf | \xbf, \Wbf_1 ) = \Psi_\sbf(\xbf, \Wbf_1 )$.
This viewpoint suggests a sampling scheme to approximate both the predictor output (Equation~\ref{eq:F2layers}) and the partial derivatives (Equations~\ref{eq:F2dw1} and~\ref{eq:F2dw2}), that can be framed as a variant of the REINFORCE algorithm \citep{Williams1992} (see the discussion below): We avoid computing the $2^{d_1}$ terms by resorting to a Monte Carlo approximation of the sum. Given an input $\xbf$ and a sampling size $T$, the procedure goes as follows.

   \textbf{Prediction.} 
   We generate $T$ random binary vectors $Z {\eqdef} \{\sbf^t\}_{t=1}^T$  according to the $ \Psi_\sbf(\xbf, \Wbf_1 )$-distribution. This can be done by uniformly sampling $z^t_i{\in}[0,1]$, and setting
   $s^t_i {=} \sgn(\psi_1 (\xbf, \wbf_1^i){-}z^t_i)$.\\
    A stochastic approximation of $F_{\theta}(\xbf)$ is given by
    $\widehat F_{\theta}(Z)  \eqdef \frac{1}{T} \sum_{t=1}^T \Erf \left( \frac{\wbf_2 \cdot \sbf^t}{\sqrt{2d_1}} \right)$ .
    
    \textbf{Derivatives.} Note that for a given sample
    $\{\sbf^t\}_{t=1}^T$, the approximate derivatives according to $\wbf_2$ (Equation~\ref{eq:approxF2dw2} below) can be computed numerically by the automatic differentiation mechanism of deep learning frameworks while evaluating $\widehat F_{\theta}(Z)$ \citep[\eg,][]{paszke2017automatic}. 
    However, we need the following Equation~\eqref{eq:approxF2dw1} to approximate the gradient according to $\Wbf_1$ because $\partial \widehat F_\theta(Z)/\partial\wbf_1^k  =0$.
    
    \begin{align}
    \label{eq:approxF2dw1}
       \frac{\partial }{\partial\wbf_1^k} F_\theta(\xbf) &\approx\frac{\xbf}{T\,2^{\frac32}\|\xbf\|}
    \Erf'\left(\frac{\wbf_1^k \cdot \xbf}{\sqrt{2}\norm{\xbf}}\right) 
    \sum_{t=1}^T
     \frac{s^t_k}{\psi_{s^t_k} (\xbf, \wbf_1^k)} \Erf \left( \frac{\wbf_2 \cdot \sbf^t}{\sqrt{2d_1}} \right);
     \\
     \label{eq:approxF2dw2}
     \frac{\partial }{\partial\wbf_2} F_\theta(\xbf)
&\approx \frac{1}{T\sqrt{2d_1}}  \sum_{t=1}^{T}
\sbf^t \, \Erf' \left( \frac{\wbf_2 \cdot \sbf^t}{\sqrt{2d_1}} \right)
= 
\frac{\partial }{\partial\wbf_2} \widehat F_\theta(Z)\,.
    \end{align}

\textbf{Similar approaches to stochastic networks.} 
Random activation functions are commonly used in generative neural networks, and tools have been developed to train these by gradient descent (see \citet[Section 20.9]{Goodfellow-16-book} for a review). Contrary to these approaches, our analysis differs as the stochastic operations are introduced to estimate a deterministic objective. That being said, Equation~\eqref{eq:approxF2dw1} can be interpreted as a variant of REINFORCE algorithm \citep{Williams1992} to apply the back-propagation method along with discrete activation functions. Interestingly, the formulation we obtain through our PAC-Bayes objective is similar to a commonly used REINFORCE variant \citep[\eg,][]{BengioLC13,YinZ19}, where the activation function is given by a Bernoulli variable with probability of success $\sigma(a)$, where  $a$ is the neuron input, and $\sigma$ is the sigma is the \emph{sigmoid function}. The latter can be interpreted as a surrogate of our $\psi_{s_i} (\xbf, \wbf^i_1)$.

\section{Generalization to multilayer networks}\label{sec:multilayers}

\begin{figure}
   \centering
      \label{fig:bam2tree}
\begin{tikzpicture}[neuralnetwork]
    \def\layersd{15mm}
    \def\neuronsize{16pt}
    
    \node[neuron] (I-1) {$x_1$};
    \node[neuron] (I-2) {$x_2$};
    
    \node[neuron,above left=\layersd and 5mm of I-1.center,fill=H1-1-col, minimum size=\neuronsize] (H1-1) {};
    \node[neuron,fill=H1-2-col, minimum size=\neuronsize] (H1-2) {};
    \node[neuron,fill=H1-3-col, minimum size=\neuronsize] (H1-3) {};
    
    \node[neuron,above=\layersd of H1-1.center,fill=H2-1-col, minimum size=\neuronsize]  (H2-1)   {};
    \node[neuron, fill=H2-2-col, minimum size=\neuronsize] (H2-2)  {};
    \node[neuron, fill=H2-3-col, minimum size=\neuronsize] (H2-3)  {};
    
    \node[neuron,above=\layersd of H2-2.center, fill=out-col, minimum size=\neuronsize]  (O-1)   {$\hat{y}$};
    
    \foreach \i in {1,2}
    \foreach \j in {1,2,3}
    {
    	\path (I-\i) edge (H1-\j);
    }
    \foreach \i in {1,2,3}
    \foreach \j in {1,2,3}
    {
    	\path (H1-\i) edge (H2-\j);
    }
    \foreach \i in {1,2,3}
    {
    	\path (H2-\i) edge (O-1);
    }
    
    
    \node[neuron, right=10 mm of I-2.center] (TI-1-1) {$x_1$};
    \node[neuron, right=1mm of TI-1-1] (TI-1-2) {$x_2$};
    
    \foreach \i in {2,...,9}
    {
    \pgfmathsetmacro{\imoinsun}{int(\i-1)};
    \node[neuron, right=1.5mm of TI-\imoinsun-2] (TI-\i-1) {$x_1$};
    \node[neuron, right=1mm of TI-\i-1] (TI-\i-2) {$x_2$};
    }
    
    \node[neuron,above right=\layersd and 1mm of TI-2-1.center,fill=H1-2-col, minimum size=\neuronsize] (TH1-1-2) {};
    \node[neuron,left=4mm of TH1-1-2,fill=H1-1-col, minimum size=\neuronsize] (TH1-1-1) {};
    \node[neuron,right=4mm of TH1-1-2, fill=H1-3-col, minimum size=\neuronsize] (TH1-1-3) {};
    
    \node[neuron,above right=\layersd and 0.5mm of TI-5-1.center,fill=H1-2-col, minimum size=\neuronsize] (TH1-2-2) {};
    \node[neuron,left=4mm of TH1-2-2,fill=H1-1-col, minimum size=\neuronsize] (TH1-2-1) {};
    \node[neuron,right=4mm of TH1-2-2, fill=H1-3-col, minimum size=\neuronsize] (TH1-2-3) {};
    
    \node[neuron,above right=\layersd and 0mm of TI-8-1.center, fill=H1-2-col, minimum size=\neuronsize] (TH1-3-2) {};
    \node[neuron,left=4mm of TH1-3-2,fill=H1-1-col, minimum size=\neuronsize] (TH1-3-1) {};
    \node[neuron,right=4mm of TH1-3-2, fill=H1-3-col, minimum size=\neuronsize] (TH1-3-3) {};
    
    \node[neuron,above=\layersd of TH1-2-2.center,fill=H2-2-col, minimum size=\neuronsize] (TH2-2) {};
    \node[neuron,left=20mm of TH2-2,fill=H2-1-col, minimum size=\neuronsize] (TH2-1) {};
    \node[neuron,right=20mm of TH2-2, fill=H2-3-col, minimum size=\neuronsize] (TH2-3) {};
    
    \node[neuron,above=\layersd of TH2-2.center, fill=out-col, minimum size=\neuronsize]  (TO-1)   {$\hat{y}$};
    
    \foreach \i in {1,2,3}
    {
    	\path (TH2-\i) edge (TO-1);
    }
    
    \foreach \i in {1,2,3}
    {
    	\foreach \j in {1,2,3}
    	{
    		\path (TH1-\i-\j) edge (TH2-\i);
    		\pgfmathsetmacro{\k}{int((\i-1)*3+\j)};
    		\path (TI-\k-1) edge (TH1-\i-\j);
    		\path (TI-\k-2) edge (TH1-\i-\j);
    		
    	}
    }
    \draw[shorten >=1.5cm,shorten <=1.5cm,->, line width=2pt] (O-1) edge (TO-1);
\end{tikzpicture}
  \caption{Illustration of the \emph{BAM to tree architecture map} on a three layers network.}
\end{figure}
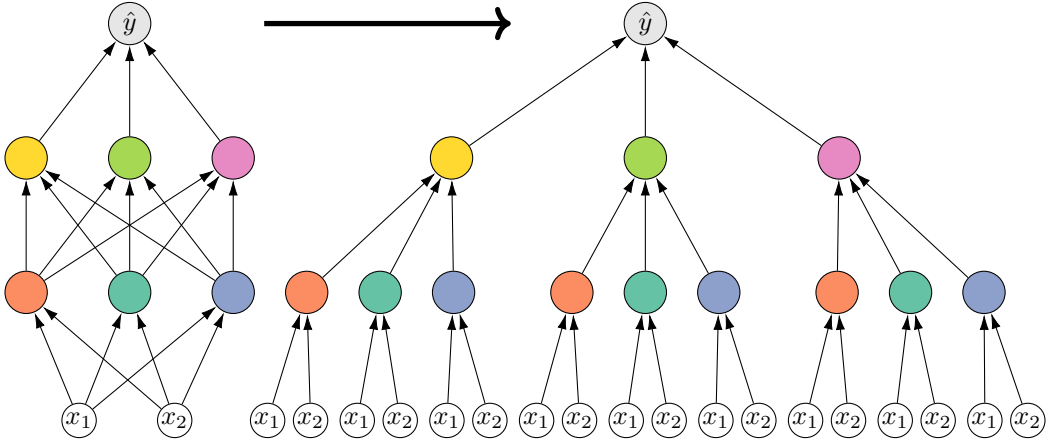

In the following, we extend the strategy introduced in Section~\ref{sec:twolayers} to BAM architectures with an arbitrary number of layers $L\in\Nds^*$ (Equation~\ref{eq:BAM}).
An apparently straightforward approach to achieve this generalization would have been to consider a Gaussian posterior distribution $\Ncal(\theta, I_D)$ over the BAM family $\{f_{\tilde\theta} | \tilde\theta\in\Rds^D\}$. However, doing so leads to a deterministic network relying on undesirable sums of 
$\prod_{k=1}^L 2^{d_k}$ elements (see Appendix~\ref{sec:withouttree} for details).
Instead, we define a mapping $f_\theta\mapsto g_{\zeta(\theta)}$ which transforms the BAM network into a computation tree, as illustrated by Figure~\ref{fig:bam2tree}.

\label{sec:tree}

\textbf{BAM to tree architecture map.}
Given a BAM network $f_\theta$ of $L$ layers with sizes $d_0,d_1,\ldots,d_L$ (reminder: $d_L{=}1$), we obtain a \emph{computation tree} by decoupling the neurons (\ie, the computation graph nodes): the tree leaves contain $\prod_{k=1}^{L} d_k$ copies of each of the $d_0$ BAM input neurons, and the tree root node corresponds to the single BAM output neuron. Each input-output path of the original BAM network becomes a path of length $L$ from one leaf to the tree root. Each tree edge has its own parameter (a real-valued scalar); the total number of edges is
$D^\dagger \eqdots \sum_{k=0}^{L-1} d^\dagger_k$, with $d^\dagger_k \eqdef \prod_{i=k}^{L} d_i$.
We define a set of tree parameters 
$\eta$ recursively according to the tree structure. From level $k$ to $k{+}1$, the tree has $d^\dagger_k$ edges. That is, each node at level $k{+}1$ has its own parameters subtree $\eta^{k+1} \eqdef  \{\eta^k_i\}_{i=0}^{d_{k}}$, where each $\eta^k_i$
is either a weight vector containing the input edges parameters (by convention, $\eta^k_0 \in \Rds^{d_{k-1}}$) or a parameter set (thus, $\eta^k_1,\ldots,\eta^k_{d_{k-1}}$ are themselves parameter subtrees).
Hence, the \emph{deepest} elements of the recursive parameters set $\eta$ are weight vectors $\eta^{1}\in\Rds^{d_0}$.
Let us now define the output tree $g_\eta(\xbf) \eqdef g_L(\xbf, \eta)$ on an input $\xbf\in\Rds^{d_0}$ as a recursive function:
\begin{align*}
g_1(\xbf, \{\wbf \})
&=
\sgn \left(\wbf\cdot\xbf\right),\\
g_{k+1}(\xbf, \underbrace{\{\wbf, \eta^k_1, \ldots, \eta^k_{d_k}\}}_{\eta^k}) 
&= \sgn\Big(\wbf\cdot
\underbrace{(g_{k}(\xbf, \eta_1), \ldots, g_{k}(\xbf, \eta_{d_{k
}}))}_{\gbf_{k}(\xbf,\eta^k)}\Big) \mbox{ for $k=1,\ldots,L{-}1$}\,.
\end{align*}

 \textbf{BAM to tree parameters map.} 
 Given BAM parameters $\theta$, 
 we denote $\theta_{1:k}\eqdef \mathrm{vec}\!\left(\{ \Wk\}_{i=1}^{k}\right)$.
 The mapping from $\theta$ into the corresponding (recursive) tree parameters set is
$\zeta(\theta)= \{\wbf_L, \zeta_1(\theta_{1:L-1}), \ldots, \zeta_{d_{L-1}}(\theta_{1:L-1})\}$, such that
$\zeta_i(\theta_{1:k}) = \{\wbf_k^i, \zeta_1(\theta_{1:k-1}), \ldots, \zeta_{d_{k-1}}(\theta_{1:k-1})\}$, 
and
$\zeta_i(\theta_{1:1}) = \{\wbf_1^i\}$. Note that the parameters tree obtained by the transformation $\zeta(\theta)$ is highly redundant, as each weight vector $\wbf_k^i$ (the $i$th line of the $\Wbf_k$ matrix from $\theta$) is replicated $d^\dagger_{k+1}$ times.
This construction is such that
$f_\theta(\xbf)= g_{\zeta(\theta)}(\xbf)$ for all $\xbf\in\Rds^{d_0}$.


\newcommand{\onek}{{1:k}}
\newcommand{\onekm}{{1:k{-}1}}
\newcommand{\ttheta}{{\tilde\theta}}
\newcommand{\ttonek}[1][k]{\ttheta_{1:#1}} 
\newcommand{\onebb}{\mathbb{1}}
\newcommand{\vki}[1][k]{\vbf_{#1}^{i}}
\newcommand{\QVki}[1][k]{Q(\vki[#1])}

\newcommand{\teta}{{\tilde\eta}}
\newcommand{\QVonek}[1][k]{Q(\teta_{1:#1})} 
\newcommand{\QVonekm}{Q(\teta_\onekm)} 

\textbf{Deterministic network.}
With a slight abuse of notation, we let $\tilde\eta \sim Q_\eta \eqdef \Ncal(\eta, I_{D^\dagger})$ denote a parameter tree of the same structure as $\eta$, where every weight is sampled \iid{} from a normal distribution. 
We denote $G_{\theta}(\xbf) \eqdef \Esp_{\teta\sim Q_{\zeta(\theta)}} g_\teta(\xbf)$, and we compute the output value of this predictor recursively. In the following, we denote $G^{_{(j)}}_{\theta_{1:k+1}}(\xbf)$ the function returning the $j$th neuron value of the layer $k{+}1$. Hence, the output of this network is $G_\theta(\xbf) = G^{_{(1)}}_{\theta_{1:L}}(\xbf)$. As such,
\begin{align} 
\nonumber
    G^{(j)}_{\theta_{1:1}}(\xbf) 
    \ =& \int_{\Rds^{d_0}} Q_{\wbf_1^j}(\vbf) \sgn(\vbf\cdot\xbf)d\vbf
    \ =\ \Erf\left(\tfrac{\wbf_1^j\cdot \xbf}{\sqrt{2} \|\xbf\|}\right),\\
G^{(j)}_{\theta_{1:k+1}}(\xbf) =& 
\sum_{\mathclap{\sbf\in\{-1,1\}^{d_{k}}}}
\Erf\left(\tfrac{\wbf_{k+1}^j\cdot\sbf}{\sqrt{2 d_{k}}}  \right) 
\Psi^{k}_\sbf(\xbf, \theta),
\mbox{ with }
\Psi^{k}_\sbf(\xbf, \theta) 
=
\prod_{i=1}^{d_{k}}
\,\left(\frac{1}{2} + \frac{1}{2} s_i \times G^{(i)}_{\theta_{1:k}}(\xbf)\right).
\label{eq:Grecursif}
\end{align}
The complete mathematical calculations leading to the above results are provided in Appendix~\ref{sec:supp_multi_pred}. The computation tree structure and the parameter mapping $\zeta(\theta)$ are crucial to obtain the recursive expression of Equation~\eqref{eq:Grecursif}. However, note that this abstract mathematical structure is never manipulated explicitly. Instead, it allows computing each hidden layer vector 
$( G^{_{(j)}}_{\theta_{1:k}}(\xbf) )_{j=1}^{d_k}$ sequentially; a summation of $2^{d_k}$ terms is required for each layer $k=1,\ldots,L{-}1$. 

\textbf{Stochastic approximation.}
Following the Section \ref{section:stochastic} sampling procedure trick for the one hidden layer network, we propose to perform a stochastic approximation of the network prediction output, by  a Monte Carlo sampling for each layer.
Likewise, we recover exact and approximate derivatives in a layer-by-layer scheme.
The related equations are given in Appendix~\ref{sec:supp_multi_deriv}.


\section{PBGNet: PAC-Bayesian SGD learning of binary activated networks}\label{sec:pbgnet}

We design an algorithm to learn the parameters $\theta\in\Rds^D$ of the predictor $G_\theta$ by minimizing a PAC-Bayesian upper bound on the generalization loss $\Lcal_\Dcal(G_\theta)$. We name our algorithm PBGNet (\textbf{P}AC-Bayesian \textbf{B}inary \textbf{G}radient \textbf{Net}work), as it is a generalization of the PBGD (PAC-Bayesian Gradient Descent) learning algorithm for linear classifiers \citep{germain2009pac} to deep binary activated neural networks. 

\textbf{Kullback-Leibler regularization.} The computation of a PAC-Bayesian bound value relies on two key elements: the empirical loss on the training set and the Kullback-Leibler divergence between the prior and the posterior. Sections~\ref{sec:twolayers} and~\ref{sec:multilayers} present exact computation and approximation schemes for the empirical loss $\widehat\Lcal_S(G_\theta)$ (which is equal to $\widehat\Lcal_S(F_\theta)$ when $L{=}2$). Equation~\eqref{eq:nonceKLnestpaspourlevieilhomme} introduces the $\KL$-divergence associated to the parameter maps of Section~\ref{sec:tree}. We use the shortcut notation $\Kcal (\theta, \mu)$ to refer to the divergence between two multivariate Gaussians of $D^\dagger$ dimensions, corresponding to learned parameters 
$\theta = \vec\! \left(\{ \Wk\}_{k=1}^{L}\right)$ and prior parameters
$\mu = \vec\! \left(\{ \Ubf_k\}_{k=1}^{L}\right)$.
\begin{equation} \label{eq:nonceKLnestpaspourlevieilhomme}
    \Kcal (\theta, \mu) \eqdef \KL\Big(Q_{\zeta(\theta)}\,\big\|\,P_{\zeta(\mu)}\Big)
    =
    \frac{1}{2} \left( \|\wbf_L-\ubf_L\|^2 + \sum_{k=1}^{L-1} 
    d^\dagger_{k+1}
    \big\|\Wbf_k-\Ubf_k\big\|^2_F\right),
\end{equation}
where  the factors $d^\dagger_{k+1}= \prod_{i=k+1}^{L} d_i$ are due to the redundancy introduced by transformation $\zeta(\cdot)$.
This has the effect of penalizing more the weights on the first layers. It might have a considerable influence on the bound value for very deep networks. On the other hand, we observe that this is consistent with the \emph{fine-tuning} practice performed when training deep neural networks for a transfer learning task: prior parameters are learned on a first dataset, and the posterior weights are learned by adjusting the last layer weights on a second dataset  \citep[see][]{bengio09,yosinski14}.

\textbf{Bound minimization.}
PBGNet minimizes the bound of 
 Theorem~\ref{thm:pbseeger} (rephrased as Equation~\ref{eq:bnd1}). However, this is done indirectly by minimizing a variation on Theorem~\ref{thm:catoni} and used in a deep learning context by \citet{zhou2018nonvacuous} (Equation~\ref{eq:bnd2}).
Theorem~\ref{thm:bnd1to2} links both results (proof in Appendix~\ref{supp:proofthm}).
\begin{thm}\label{thm:bnd1to2}
Given prior parameters $\mu\in\Rds^D$, with probability at least $1-\delta$ over $S\sim \Dcal^{n}$, we have for all $\theta$ on $\Rds^D$\,:
\begin{align} \label{eq:bnd1}
   \Lcal_\Dcal (G_\theta) \, &\leq
         \sup_{0\leq p \leq 1} \left\{p :  \kl(\widehat\Lcal_S (G_\theta) \| p ) \leq  \frac1n [\Kcal (\theta, \mu)+\ln\tfrac{2\sqrt{n}}{\delta}]\right\}\\
   &= 
    \label{eq:bnd2}
\inf_{C>0} \left\{\tfrac{1}{1-e^{-C}}\left(
     1-\exp\left(-C \, \widehat\Lcal_S (G_\theta) - \frac1n [\Kcal (\theta, \mu)+\ln\tfrac{2\sqrt{n}}{\delta}]\right)
     \right)   \right\}.
\end{align}
\end{thm}
We use stochastic gradient descent (SGD) as the optimization procedure to minimize Equation~\eqref{eq:bnd2} with respect to $\theta$ and $C$. It optimizes the same trade-off as in Equation~\eqref{eq:pbgd3}, but choosing the $C$ value which minimizes the bound.\footnote{We also note that our training objective can be seen as a generalized Bayesian inference one \citep{knoblauch2019}, where the tradeoff between the loss and the KL divergence is given by the PAC-Bayes bound.}
The originality of our SGD approach is that not only do we induce gradient randomness by selecting \emph{mini-batches} among the training set $S$, we also approximate the loss gradient by sampling $T$ elements for the combinatorial sum at each layer.
Our experiments show that, for some learning problems, reducing the sample size of the Monte Carlo approximation can be beneficial to the stochastic gradient descent.
Thus the sample size value $T$ has an influence on the cost function space exploration during the training procedure (see Figure \ref{fig:sample_effect} in the appendix). 
Hence, we consider $T$ as a PBGNet hyperparameter.



\section{Numerical experiments}\label{sec:xp}


Experiments were conducted on six binary classification datasets, described in Appendix~\ref{sec:supp_exp}.

\textbf{Learning algorithms.}
In order to get insights on the trade-offs promoted by the PAC-Bayes bound minimization, we compared PBGNet to variants focusing on empirical loss minimization. We train the models using multiple network architectures (depth and layer size) and hyperparameter choices. The objective is to evaluate the efficiency of our PAC-Bayesian framework both as a learning algorithm design tool and a model selection criterion. 
For all methods, the network parameters are trained using the Adam optimizer \citep{kingma2014adam}.
Early stopping is used to interrupt the training when the cost function value is not improved for 20 consecutive epochs. 
Network architectures explored range from 1 to 3 hidden layers ($L$) and a hidden size $h\in\{10, 50, 100\}$ ($d_k=h$ for $1\leq k < L$). 
Unless otherwise specified, the same randomly initialized parameters are used as a prior in the bound and as a starting point for SGD optimization \citep[as in][]{Karolina2017}.
Also, for all models except MLP, we select the binary activation sampling size $T$ in a range  going from 10 to 10000.
More details about the experimental setting are given in Appendix~\ref{sec:supp_exp}. 

\newcommand{\paralgo}[1]{\vspace{-1.1mm}\textit{#1}.\ }

\paralgo{MLP} We compare to a 
standard 
network with $\tanh$ activation, as this activation resembles the $\Erf$ function of PBGNet.
We optimize the linear loss as the cost function and use 20\% of training data as validation for hyperparameters selection.
A weight decay parameter $\rho$ is selected between $0$ and $10^{-4}$. Using weight decay corresponds to adding an $L2$ regularizer $\frac\rho2\|\theta\|^2$ to the cost function, but contrary to the regularizer of Equation~\eqref{eq:nonceKLnestpaspourlevieilhomme} promoted by PBGNet, this regularization is uniform for all layers. 

\paralgo{PBGNet$_{\ell}$} This variant minimizes the empirical loss $\widehat\Lcal(G_\theta)$, with an $L2$ regularization term $\frac\rho2\|\theta\|^2$. The corresponding weight decay $\rho$, as well as other hyperparameters, are selected using a validation set, exactly as the MLP does. 
The bound expression is not involved in the learning process and is computed on the model selected by the validation set technique. 

\paralgo{PBGNet$_{\ell\text{-bnd}}$}
Again, the empirical loss $\widehat\Lcal(G_\theta)$ with an $L2$ regularization term $\frac\rho2\|\theta\|^2$ is minimized. However, only the weight decay hyperparameter $\rho$ is selected on the validation set,
the other ones are selected by the bound.
This method is motivated by an empirical observation: our PAC-Bayesian bound is a great model selection tool for most hyperparameters, except the weight decay term.


\paralgo{PBGNet} As described in Section \ref{sec:pbgnet}, the generalization bound is directly optimized as the cost function during the learning procedure and used solely for hyperparameters selection:
no validation set is needed and all training data $S$ are exploited for learning.

\paralgo{PBGNet$_{\text{pre}}$}
We also explore the possibility of using a part of the training data as a pre-training step.
To do so, we split the training set into two halves.
First, we minimize the empirical loss for a fixed number of 20 epochs on the first 50\% of the training set.
Then, we use the learned parameters as initialization and prior for PBGNet and learn on the second 50\% of the training set.


\textbf{Analysis.}
Results are summarized in Table \ref{tab:main_results}, which highlights the strengths and weaknesses of the models.
Both MLP and PBGNet$_{\ell}$ obtain competitive error scores but lack generalization guarantees.
By introducing the bound value in the model selection process, even with the linear loss as the cost function, PBGNet$_{\ell\text{-bnd}}$ yields non-vacuous generalization bound values although with an increase in error scores.
Using the bound expression for the cost function in PBGNet improves bound values while keeping similar performances.
The Ads dataset is a remarkable exception where the small amount of training examples seems to radically constrain the network in the learning process as it hinders the KL divergence growth in the bound expression.
With an informative prior from pre-training, PBGNet$_\text{pre}$ is able to recover competitive error scores while offering tight generalization guarantees. All selected hyperparameters are presented in the appendix (Table~\ref{tab:selected_models_overview}).

A notable observation is the impact of the bound exploitation for model selection on the train-test error gap.
Indeed, PBGNet$_{\ell\text{-bnd}}$, PBGNet and PBGNet$_\text{pre}$ display test errors closer to their train errors, as compared to MLP and PBGNet$_{\ell}$.
This behavior is more noticeable as the dataset size grows and suggests potential robustness to overfitting when the bound is involved in the learning process.
\begin{table}[t]
  \caption{Experiment results for the considered models on the binary classification datasets: error rates on the train and test sets (E$_{S}$ and E$_{T}$), and generalization bounds on the linear loss $\Lcal_\Dcal$ (Bnd).
  The PAC-Bayesian bounds hold with probability $0.95$. Bound values for PBGNet$_{\ell}$ are trivial, excepted Adult with a bound value of 0.606, and are thus not reported.
  A visual representation of this table is presented in the appendix (Figure~\ref{fig:res_sum_viz}).}
  \centering
  \setlength{\tabcolsep}{2.5pt}
  {\normalsize
  \begin{tabular}{lrrrrrrrrrrrrr}
    \toprule
    \multirow{2}{*}[-3pt]{Dataset} &  \multicolumn{2}{c}{MLP} & \multicolumn{2}{c}{PBGNet$_{\ell}$} & \multicolumn{3}{c}{PBGNet$_{\ell\text{-bnd}}$} & \multicolumn{3}{c}{PBGNet} & 
    \multicolumn{3}{c}{PBGNet$_{\text{pre}}$}\\
    \cmidrule(lr){2-3} \cmidrule(lr){4-5} \cmidrule(lr){6-8} \cmidrule(lr){9-11}\cmidrule(lr){12-14}
     & E$_{S}$ & E$_{T}$ & E$_{S}$ & E$_{T}$ & E$_{S}$ & E$_{T}$ & Bnd & E$_{S}$ & E$_{T}$ & Bnd & E$_{S}$ & E$_{T}$ & Bnd\\
    \midrule
    ads & 0.021 & 0.035 & 0.018 & \textbf{0.030} & 0.028 & 0.047 & 0.763 & 0.131 & 0.168 & 0.205 & 0.033 & 0.033 & 0.060\\
    adult & 0.137 & 0.152 & 0.133 & \textbf{0.149} & 0.147 & 0.155 & 0.281 & 0.154 & 0.163 & 0.214 & 0.149 & 0.154 & 0.164\\
    mnist17 & 0.002 & \textbf{0.004} & 0.003 & \textbf{0.004} & 0.004 & 0.006 & 0.096 & 0.005 & 0.007 & 0.040 & 0.004 & \textbf{0.004} & 0.010\\
    mnist49 & 0.004 & \textbf{0.013} & 0.003 & 0.018 & 0.029 & 0.035 & 0.311 & 0.035 & 0.040 & 0.139 & 0.016 & 0.017 & 0.028\\
    mnist56 & 0.004 & 0.013 & 0.003 & 0.011 & 0.022 & 0.024 & 0.172 & 0.022 & 0.025 & 0.090 & 0.009 & \textbf{0.009} & 0.018\\
    mnistLH & 0.006 & \textbf{0.018} & 0.004 & 0.019 & 0.046 & 0.051 & 0.311 & 0.049 & 0.052 & 0.160 & 0.026 & 0.027 & 0.033\\
    \bottomrule
\end{tabular}}
\label{tab:main_results}
\end{table}

\section{Conclusion and perspectives}\label{sec:conclusion}

We made theoretical and algorithmic contributions towards a better understanding of generalization abilities of binary activated multilayer networks, using PAC-Bayes. Note that the computational complexity of a learning epoch of PBGNet is higher than the cost induced in \emph{binary neural networks} \citep{bengio09,hubara2016binarized,DBLP:conf/nips/SoudryHM14,hubara2017quantized}. 
Indeed, we focus on the optimization of the generalization guarantee more than computational complexity. Although we also propose a sampling scheme that considerably reduces the learning time required by our method, achieving a nontrivial tradeoff.

 
We intend to investigate how we could leverage the bound to learn suitable priors for PBGNet. Or equivalently, finding (from the bound point of view) the best network architecture. 
We also plan to extend our analysis to multiclass and multilabel prediction, and convolutional networks. 
We believe that this line of work 
is part of a necessary effort to give rise to a better understanding of the behavior of deep neural networks. 

\newpage
\subsubsection*{Acknowledgments}
We would like to thank Mario Marchand for the insight leading to the Theorem~\ref{thm:bnd1to2}, Gabriel Dubé and Jean-Samuel Leboeuf for their input on the theoretical aspects, Frédérik Paradis for his help with the implementation, and Robert Gower for his insightful comments.
This work was supported in part by the French Project APRIORI ANR-18-CE23-0015, in part by NSERC and in part by Intact Financial Corporation.
We gratefully acknowledge the support of NVIDIA Corporation with the donation of Titan Xp GPUs used
for this research.

\bibliographystyle{plainnat}
\bibliography{references}

\clearpage

\appendix

\section{Supplementary Material}

 \subsection{From the \emph{sign} activation to the \emph{erf} function} 
 \label{sec:sgn2erf}
 
 For completeness, we present the detailed derivation of Equation~\eqref{eq:erfactivated}. This result appears namely in \citet{langford-02,langford-05,germain2009pac}.
 
 Given $\xbf\in\Rds^d$, we have
\begin{align*} 
F_\wbf(\xbf)
&=
    \Esp_{\vbf \sim \Ncal(\wbf, \mathrm{I})} \sgn(\vbf \cdot \xbf)\\
    &= \int_{\Rds^{d}} \sgn(\vbf \cdot \xbf) \left(\frac{1}{\sqrt{2\pi}}\right)^{d}e^{-\frac{1}{2}\norm{\vbf - \wbf}^{2}}d\vbf\\
    &= \int_{\Rds^{d}} \left( \mathbb{1}\left[\vbf \cdot \xbf > 0\right] - \mathbb{1}\left[\vbf \cdot \xbf < 0\right] \right) \left(\frac{1}{\sqrt{2\pi}}\right)^{d}e^{-\frac{1}{2}\norm{\vbf - \wbf}^{2}}d\vbf\\
    &= \left(\frac{1}{\sqrt{2\pi}}\right)^{d}\int_{\Rds^{d}} \mathbb{1}\left[\vbf \cdot \xbf > 0\right]e^{-\frac{1}{2}\norm{\vbf - \wbf}^{2}}d\vbf - \left(\frac{1}{\sqrt{2\pi}}\right)^{d}\int_{\Rds^{d}} \mathbb{1}\left[\vbf \cdot \xbf < 0\right]e^{-\frac{1}{2}\norm{\vbf - \wbf}^{2}}d\vbf\,.
\end{align*}
Without loss of generality, let us consider a vector basis where $\frac{\xbf}{\norm{\xbf}}$ is the first coordinate. In this basis, the first elements of the vectors $\vbf = (v_1, v_2, \dots, v_d)$ and $\wbf  = (w_1, w_2, \dots, w_d)$ are
\begin{align*}
    v_1 = \frac{\vbf \cdot \xbf}{\norm{\xbf}}, &&  w_1 = \frac{\wbf \cdot \xbf}{\norm{\xbf}}\,.
\end{align*}
Hence, $\vbf \cdot \xbf = v_1 \cdot \norm{\xbf}$ with $\norm{\xbf} > 0$. Looking at the left side of the subtraction from the previous equation, we thus have
\begin{align*} 
    \left(\frac{1}{\sqrt{2\pi}}\right)^{d}\int_{\Rds^{d}} & \mathbb{1}\left[\vbf \cdot \xbf > 0\right]e^{-\frac{1}{2}\norm{\vbf - \wbf}^{2}}d\vbf\\
    &= \int_{\Rds} \mathbb{1}\left[v_1 > 0\right]\frac{1}{\sqrt{2\pi}}e^{-\frac{1}{2}(v_1 - w_1)^{2}}\left[\int_{\Rds^{d-1}} \left(\frac{1}{\sqrt{2\pi}}\right)^{d-1}e^{-\frac{1}{2}\norm{\vbf_{2:d} - \wbf_{2:d}}^{2}}d\vbf_{2:d}\right]dv_1\\
    &=\frac{1}{\sqrt{2\pi}} \int_{-\infty}^{\infty} \mathbb{1}\left[t > -w_1\right]e^{-\frac{1}{2}t^{2}}dt\,,
\end{align*}
with $t \eqdef v_1 - w_1$. Hence,
\begin{align*} 
    \Esp_{\vbf \sim \Ncal(\wbf, \mathrm{I})} \sgn(\vbf \cdot \xbf)
    &= \frac{1}{\sqrt{2\pi}} \int_{-\infty}^{\infty} \mathbb{1}\left[t > -w_1\right]e^{-\frac{1}{2}t^{2}}dt - \frac{1}{\sqrt{2\pi}} \int_{-\infty}^{\infty} \mathbb{1}\left[t < -w_1\right]e^{-\frac{1}{2}t^{2}}dt\\
     &= \frac{1}{\sqrt{2\pi}} \int_{-w_1}^{\infty} e^{-\frac{1}{2}t^{2}}dt - \frac{1}{\sqrt{2\pi}} \int_{-\infty}^{-w_1} e^{-\frac{1}{2}t^{2}}dt\\
     &= \frac{1}{2} + \frac{1}{\sqrt{2\pi}} \int_{0}^{w_1} e^{-\frac{1}{2}t^{2}}dt - \frac{1}{2} + \frac{1}{\sqrt{2\pi}} \int_{0}^{w_1} e^{-\frac{1}{2}t^{2}}dt\\
     &= \frac{\sqrt{2}}{\sqrt{\pi}} \int_{0}^{w_1}
     e^{-\frac{1}{2}t^{2}}dt\\
     &= \frac{2}{\sqrt{\pi}} \int_{0}^{\frac{w_1}{\sqrt{2}}}
     e^{-u^{2}}du \quad\mbox{ with } u = \tfrac{t}{\sqrt 2}\\
     &= \Erf\left(\frac{w_1}{\sqrt{2}}\right)\\
     &= \Erf\left(\frac{\wbf \cdot \xbf}{\sqrt{2}\norm{\xbf}}\right),
\end{align*}
where $\Erf(\cdot)$ is the Gauss error function defined as $\Erf(x) = \frac{2}{\sqrt{\pi}}\int_{0}^{x}e^{-t^{2}}dt$.

\begin{figure}
  \centering
  \includegraphics[trim={0.5cm 0.5cm 0.3cm 0.5cm},clip,width=0.7\linewidth]{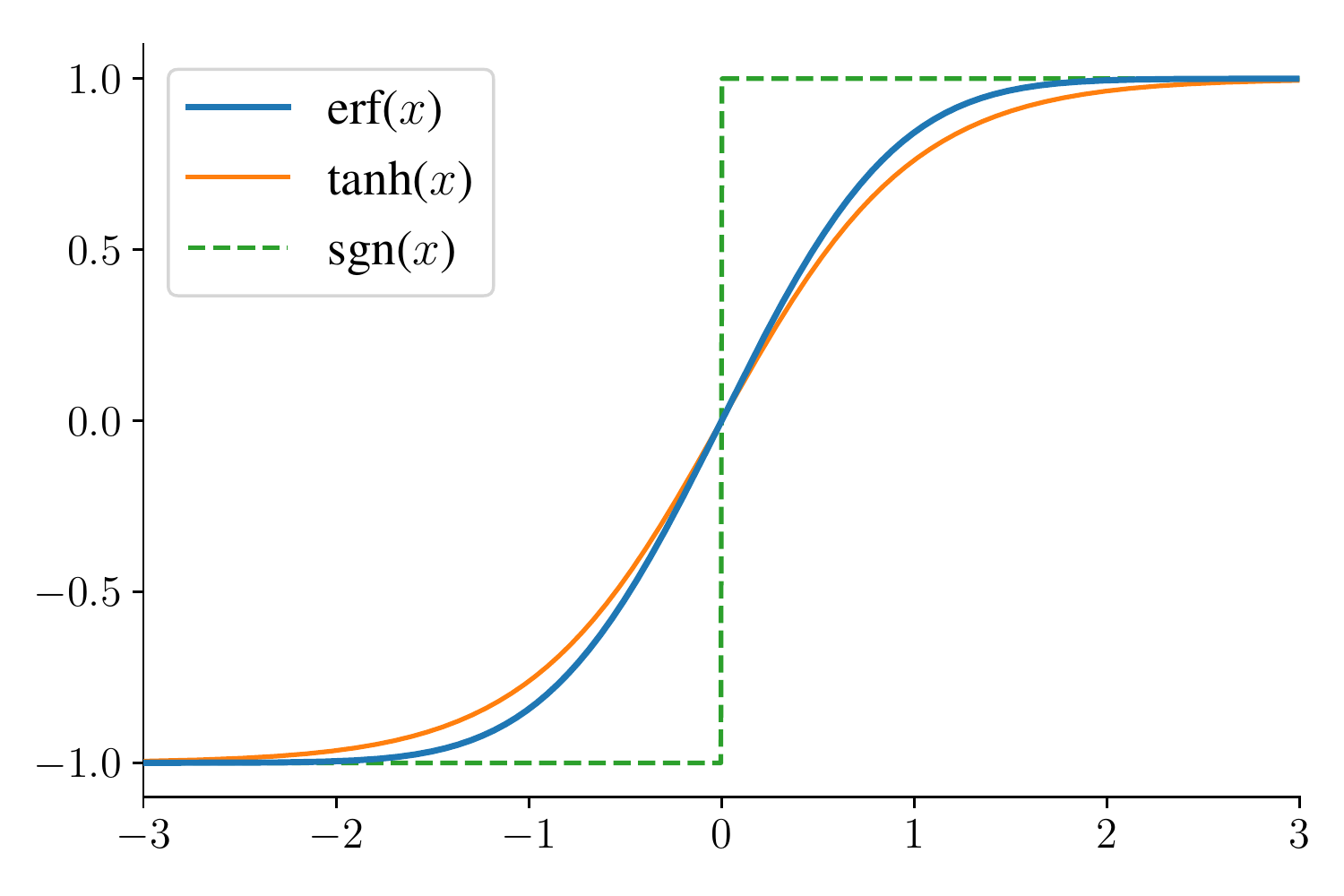}
  \caption{Visual comparison between the $\Erf$,  $\tanh$ and $\sgn$ activation functions.
  }
  \label{fig:act_func}
\end{figure}

\subsection{Aggregation of multilayer networks without the tree architecture}
\label{sec:withouttree}

To understand the benefit of the {\bf BAM to tree architecture map} introduced in Section~\ref{sec:multilayers}, let us compute the prediction function of an aggregation of two hidden layers networks following the same strategy as for the single hidden layer case (see Subsection~\ref{sec:derterministic_network}).

The two hidden layer network is parameterized by weights $\theta=\vec(\{\Wbf_1, \Wbf_2, \wbf_3\})$, with $\Wbf_1\in\Rds^{d_1\times d_0}$, $\Wbf_2\in\Rds^{d_2\times d_1}$ and $\wbf_3\in\Rds^{d_2}$. Given an input $\xbf\in\Rds^{d_0}$, the output of the network is
\begin{equation} \label{eq:3BAM}
    f_{\theta}(\xbf) = \sgn\big(\wbf_3 \cdot \sgn(\Wbf_2 \cdot \sgn(\Wbf_1 \xbf))\big)\,.
\end{equation}
We seek to compute $F_\theta(\xbf) = \Esp_{\tilde\theta \sim Q_\theta} f_{\tilde\theta} ( \xbf)$ with $\tilde\theta = \vec(\{\Vbf_1, \Vbf_2, \vbf_3\})$ and $Q_\theta = \Ncal(\theta, I_D)$. First, we need to decompose the probability of each $\tilde\theta \sim Q_\theta$ as
$Q_\theta(\tilde\theta) {=} Q_{1}(\Vbf_1) Q_{2}(\Vbf_2) Q_{3}(\vbf_3)$, with $Q_{1}{=} \Ncal(\Wbf_1, I_{d_0 d_1})$, $Q_{2}{=}\Ncal(\Wbf_2, I_{d_1 d_2})$ and  $Q_{3}{=}\Ncal(\wbf_3, I_{d_2})$.
{\allowdisplaybreaks
\begin{align*}
    F_\theta(\xbf) =& \Esp_{\tilde\theta\sim Q_\theta} f_{\tilde\theta} (\xbf)\\
    = & \int_{\mathrlap{\Rds^{d_1\times d_0}}} \  Q_1 (\Vbf_1) \int_{\mathrlap{\Rds^{d_2\times d_1}}} \ Q_2 (\Vbf_2) \int_{\mathrlap{\Rds^{d_2}}} \ Q_3 (\vbf_3) \sgn(\vbf_3 \cdot \sgn( \Vbf_2 \sgn( \Vbf_1 \xbf))) d\vbf_3 d\Vbf2 d\Vbf_1
    \nonumber \\
    =&  \int_{\mathrlap{\Rds^{d_1\times d_0}}} \  Q_1 (\Vbf_1) \int_{\mathrlap{\Rds^{d_2\times d_1}}} \ Q_2 (\Vbf_2) \, \Erf \left( \tfrac{\wbf_3 \cdot \sgn( \Vbf_2 \sgn( \Vbf_1 \xbf))}{\sqrt{2} \| \sgn( \Vbf_2 \sgn( \Vbf_1 \xbf))\|} \right) d\Vbf_2 d\Vbf_1\\
    =& \sum_{\mathclap{\tbf\in\{-1,1\}^{d_2}}} \Erf \left( \tfrac{\wbf_3 \cdot \tbf}{\sqrt{2d_2}} \right)  \int_{\mathrlap{\Rds^{d_1\times d_0}}} \  Q_1 (\Vbf_1) \int_{\mathrlap{\Rds^{d_2\times d_1}}} \ Q_2 (\Vbf_2) \  \mathbb{1}[\tbf=\sgn( \Vbf_2 \sgn( \Vbf_1 \xbf))]\,  d\Vbf_2d\Vbf_1 \\
    =&\sum_{\mathclap{\tbf\in\{-1,1\}^{d_2}}} \Erf \left( \tfrac{\wbf_3 \cdot \tbf}{\sqrt{2d_2}} \right)  \int_{\mathrlap{\Rds^{d_1\times d_0}}} \  Q_1 (\Vbf_1) \prod_{i = 1}^{d_2} \int_{\mathrlap{\Rds^{d_2\times d_1}}} \ Q_2^i (\vbf_2^i) \  \mathbb{1}[t_i \sgn( \vbf_2^i \sgn( \Vbf_1 \xbf)) > 0] \, d\vbf_2^i d\Vbf_1 \\
    =&\sum_{\mathclap{\tbf\in\{-1,1\}^{d_2}}} \Erf \left( \tfrac{\wbf_3 \cdot \tbf}{\sqrt{2d_2}} \right) \sum_{\mathclap{\sbf\in\{-1,1\}^{d_1}}} \quad\,\,\, \int_{\mathrlap{\Rds^{d_1\times d_0}}} \  Q_1 (\Vbf_1) \mathbb{1}[\sbf = \sgn( \Vbf_1 \xbf)] d\Vbf_1 \prod_{i = 1}^{d_2} \left[\frac{1}{2} + \frac{t_i}{2}\,\Erf\!\left(\frac{\wbf_2^i \cdot \sbf}{\sqrt{2d_1}}\right)  \right] \\
    =&\sum_{\mathclap{\tbf\in\{-1,1\}^{d_2}}} \Erf \left( \tfrac{\wbf_3 \cdot \tbf}{\sqrt{2d_2}} \right) \sum_{\mathclap{\sbf\in\{-1,1\}^{d_1}}} \quad\,\,\, 
    \underbrace{
    \prod_{j = 1}^{d_1} \left[\frac{1}{2} + \frac{s_j}{2}\,\Erf\!\left(\frac{\wbf_1^j \cdot \xbf}{\sqrt{2}\norm{\xbf}}\right)  \right]
    }_{\Psi_{\sbf} (\xbf, \Wbf_1)}
    \underbrace{
    \prod_{i = 1}^{d_2} \left[\frac{1}{2} + \frac{t_i}{2}\,\Erf\!\left(\frac{\wbf_2^i \cdot \sbf}{\sqrt{2d_1}}\right)  \right] 
    }_{\Psi_{\tbf} (\sbf, \Wbf_2)}
\end{align*}
}%
For each combination of first layer activation values $\sbf\in\{-1,1\}^{d_1}$ and second layer activation values $\tbf\in\{-1,1\}^{d_2}$, one needs to compute its probability $\Psi_{\sbf} (\xbf, \Wbf_1) \Psi_{\tbf} (\sbf, \Wbf_2)$. This leads to a summation of $2^{d_1} \times 2^{d_2}$ terms. Instead, the layer by layer computation obtained by our tree mapping trick implies $2^{d_1} + 2^{d_2}$ terms. This strategy enables a forward computation similar to traditional neural network, as each hidden layer values relies solely on the values of the previous layer.

\subsection{Prediction for the multilayer case (with the proposed tree architecture map)}\label{sec:supp_multi_pred}

Details of the complete mathematical calculations leading to Equation (\ref{eq:Grecursif}) are presented below:
\begin{align*}
G^{(j)}_{\theta_{1:k+1}}(\xbf) \eqdots& 
\int  Q_{\zeta_j(\theta_{1:k+1})}(\teta)\, g_{k+1} (\xbf, \teta)  \, d\teta\\
= &
\int  Q_{\zeta(\theta_{1:k})}(\teta_1) \ldots \int  Q_{\zeta(\theta_{1:k})}(\teta_{d_{k}})
\left(\int_{\Rds^{d^{k}}} Q_{\wbf_k^j}(\vbf) \sgn [\vbf \cdot \gbf_{k}(\xbf, \teta)] d \vbf \right)
d \teta_{d_{k}} \dots d \teta_1
\\
= &
\int  Q_{\zeta(\theta_{1:k})}(\teta_1) \ldots \int  Q_{\zeta(\theta_{1:k})}(\teta_{d_{k}})
\,\Erf\left(\frac{\wbf_{k+1}^j\cdot\gbf_{k}(\xbf, \teta)}{\sqrt 2 \|\gbf_{k}(\xbf, \teta)\|}  \right) 
d \teta_{d_{k}} \dots d \teta_1
\\
= &
\sum_{\mathclap{\sbf\in\{-1,1\}^{d_{k}}}}
\Erf\left(\frac{\wbf_{k+1}^j\cdot\sbf}{\sqrt{2 d_{k}}}  \right) 
\int  Q_{\zeta(\theta_{1:k})}(\teta_1) \ldots \int  Q_{\zeta(\theta_{1:k})}(\teta_{d_{k}})
\,\onebb[\sbf=\gbf_{k}(\xbf, \teta)]
d \teta_{d_{k}} \cdots d \teta_1
\\
= &
\sum_{\mathclap{\sbf\in\{-1,1\}^{d_{k}}}}
\Erf\left(\frac{\wbf_{k+1}^j\cdot\sbf}{\sqrt{2 d_{k}}}  \right) 
\prod_{i=1}^{d_{k}}
\int  Q_{\zeta(\theta_{1:k})}(\teta_i) 
\,\onebb[s_i= g_{k}(\xbf, \teta_i)]
d \teta_{i} 
\\
= &
\sum_{\mathclap{\sbf\in\{-1,1\}^{d_{k}}}}
\Erf\left(\frac{\wbf^j_{k+1}\cdot\sbf}{\sqrt{2 d_{k}}}  \right) 
\prod_{i=1}^{d_{k}}
\int  Q_{\zeta(\theta_{1:k})}(\teta_i) 
\,\left(\frac{1}{2} + \frac{s_i}{2} g_{k}(\xbf, \teta_i)\right)
d \teta_{i} 
\\
= &
\sum_{\mathclap{\sbf\in\{-1,1\}^{d_{k}}}}
\Erf\left(\frac{\wbf^j_{k+1}\cdot\sbf}{\sqrt{2 d_{k}}}  \right) 
\prod_{i=1}^{d_{k}}
\,\left(\frac{1}{2} + \frac{s_i}{2} \int  Q_{\zeta(\theta_{1:k})}(\teta_i) g_{k}(\xbf, \teta_i)d \teta_{i} \right)
\\
= &
\sum_{\mathclap{\sbf\in\{-1,1\}^{d_{k}}}}
\Erf\left(\frac{\wbf^j_{k+1}\cdot\sbf}{\sqrt{2 d_{k}}}  \right)
\underbrace{
\prod_{i=1}^{d_{k}}
\,\left(\frac{1}{2} + \frac{1}{2} s_i \times G^{(i)}_{\theta_{1:k}}(\xbf) \right)
}_{\Psi^{k}_\sbf(\xbf, \eta)}
.\end{align*}
Moreover,
\begin{align*}
\Psi^{k}_\sbf(\xbf, \theta) 
=
\prod_{i=1}^{d_{k}}
\underbrace{
\,\left(\frac{1}{2} + \frac{1}{2} s_i \times G^{(i)}_{\theta_{1:k}}(\xbf)\right)
}_{\psi^{k}_{s_i}(\xbf, \theta)}\,. 
\end{align*}
Base case:
\begin{align*}
    G^{(j)}_{\theta_{1:1}}(\xbf) &= \Esp_{\teta \sim \Ncal(\zeta_j(\theta_{1:1}),I)}  g_1 (\xbf, \teta)\\
    &= \int_{\Rds^{d_0}} Q_{\wbf_1^j}(\vbf) \sgn(\vbf\cdot\xbf)d\vbf\\
    &= \Erf\left(\frac{\wbf_1^j\cdot \xbf}{\sqrt{2} \|\xbf\|}\right).
\end{align*}

\newpage
\subsection{Derivatives of the multilayer case (with the proposed tree architecture map)}\label{sec:supp_multi_deriv}

We first aim at computing $\frac{\partial }{\partial\wbf_{k+1}} G_{\theta_{1:k+1}}(\xbf)$.

Recall that $\wbf_k^j \in \{\wbf_k^1, \ldots, \wbf_k^{d_k}\}$ is the $j^{\rm th}$ line of $\Wbf_k$, that is the input weights of the corresponding hidden layer's neuron. 
\begin{align*}
\frac{\partial }{\partial\wbf_{k+1}^j} G^{(j)}_{\theta_{1:k+1}}(\xbf)
    =&\frac{\partial}{\partial\wbf_{k+1}^j} \quad\,\,\sum_{\mathclap{\sbf\in\{-1,1\}^{d_{k}}}}
    \Erf\left(\tfrac{\wbf_{k+1}^j\cdot\sbf}{\sqrt{2 d_{k}}}  \right) 
    \Psi^{k}_\sbf(\xbf, \theta)\\
    =& \sum_{\mathclap{\sbf\in\{-1,1\}^{d_{k}}}} \quad\frac{\sbf}{\sqrt{2 d_{k}}} \Erf'\left(\tfrac{\wbf_{k+1}^j\cdot\sbf}{\sqrt{2 d_{k}}}  \right)
    \Psi^{k}_\sbf(\xbf, \theta)\,.
\end{align*}
The base case of the recursion is
\begin{align*}
\frac{\partial }{\partial\wbf_1^j} G^{(j)}_{\theta_{1:1}}(\xbf) 
    =&\frac{\partial}{\partial\wbf_1^j} \Erf\left(\frac{\wbf_1^j\cdot \xbf}{\sqrt{2} \|\xbf\|}\right)\\
    =& \frac{\xbf}{\sqrt{2}\|\xbf\|}\Erf'\left(\frac{\wbf_1^j\cdot \xbf}{\sqrt{2} \|\xbf\|}\right).
\end{align*}

In order to propagate the error through the layers, we also need to compute for $k > 1$:
\begin{align*}
\frac{\partial }{\partial G^{(l)}_{\theta_{1:k}}} G^{(j)}_{\theta_{1:k+1}}
    =&\frac{\partial}{\partial G^{(l)}_{\theta_{1:k}}} \quad\,\,\sum_{\mathclap{\sbf\in\{-1,1\}^{d_{k}}}}
    \Erf\left(\frac{\wbf^j_{k+1}\cdot\sbf}{\sqrt{2 d_{k}}}  \right) 
    \prod_{i=1}^{d_{k}}
    \,\left(\frac{1}{2} + \frac{1}{2} s_i \times G^{(i)}_{\theta_{1:k}} \right)\\
    =& \sum_{\mathclap{\sbf\in\{-1,1\}^{d_{k}}}}
    \Erf\left(\frac{\wbf^j_{k+1}\cdot\sbf}{\sqrt{2 d_{k}}}\right) \left[\frac{\Psi_\sbf^{k} (\xbf, \theta)}{\psi^k_{s_l} (\xbf, \theta)}
    \right] 
   \frac{\partial}{\partial G^{(l)}_{\theta_{1:k}}} \psi^k_{s_l} (\xbf, \theta)\\
   =& \sum_{\mathclap{\sbf\in\{-1,1\}^{d_{k}}}}
    \Erf\left(\frac{\wbf^j_{k+1}\cdot\sbf}{\sqrt{2 d_{k}}}\right) \left[\frac{s_l\Psi_\sbf^{k} (\xbf, \theta)}{2 \psi^k_{s_l} (\xbf, \theta)}
    \right].
\end{align*}

Thus, we can compute
\begin{equation*}
    \frac{\partial \widehat\Lcal_S \left(G^{(j)}_{\theta_{1:k+1}}(\xbf)\right)}{\partial\wbf_{k}^{j}} = \sum_{l}\frac{\partial \widehat\Lcal_S \left(G^{(j)}_{\theta_{1:k+1}}(\xbf)\right)}{\partial G^{(l)}_{\theta_{1:k+1}}}\frac{\partial G^{(l)}_{\theta_{1:k+1}} }{\partial G^{(j)}_{\theta_{1:k}}}\frac{\partial G^{(j)}_{\theta_{1:k}}}{\partial\wbf_{k}^{j}}\,.
\end{equation*}

\newpage
\subsection{Proof of Theorem~\ref{thm:bnd1to2}}
\label{supp:proofthm}  

\begin{lem}[\citet{germain2009pac}, Proposition~2.1; \citet{lacasse-these}, Proposition 6.2.2]
\label{lem:supC=kl}~\\
For any $0\leq q\leq p <1$, we have
\begin{equation*} \label{eq:supC=kl}
    \sup_{C>0}   \Big[\Delta(C, q, p)\Big] = \kl(q \|p)\,,
\end{equation*}
with 
\begin{align} \label{eq:DeltaC}
  \Delta(C, q,p)  \, \eqdef \, -\ln(1-p(1-e^{-C}))-Cq\,.  
\end{align}

\end{lem}
\begin{proof}
For $0\leq q, p <1$, $\Delta(C, q,p)$ is concave in $C$ and the maximum is $c_0= -\ln\left( \frac{qp-p}{qp-q}\right)$. Moreover, $\Delta(c_0, q,p) = \kl(q \|p)$.
\end{proof}

\textbf{Theorem~\ref{thm:bnd1to2}.} 
{\it
Given prior parameters $\mu\in\Rds^D$, with probability at least $1-\delta$ over $S\sim \Dcal^{n}$, we have for all $\theta$ on $\Rds^D$\,:
\begin{align*}
   \Lcal_\Dcal (G_\theta) \, &\leq
         \sup_{0\leq p \leq 1} \left\{p :  \kl(\widehat\Lcal_S (G_\theta) \| p ) \leq  \frac1n [\Kcal (\theta, \mu)+\ln\tfrac{2\sqrt{n}}{\delta}]\right\}\\
   &= 
\inf_{C>0} \left\{ \tfrac{1}{1-e^{-C}}\left(
     1-\exp\left(-C \, \widehat\Lcal_S (G_\theta) - \frac1n [\Kcal (\theta, \mu)+\ln\tfrac{2\sqrt{n}}{\delta}]\right)
     \right)   \right\}.
\end{align*}
}
\begin{proof}
In the following, we denote $\xi \eqdef \frac1n [\Kcal (\theta, \mu)+\ln\tfrac{2\sqrt{n}}{\delta}] $ and we assume 
$0<\xi<\infty$.
Let us define
\begin{equation}\label{eq:pstar}
    p^* \eqdef \sup_{0\leq p \leq 1} \left\{p :  \kl(\widehat\Lcal_S (G_\theta) \| p ) \leq  \xi \right\}.
\end{equation}
First, by a straightforward rewriting of Theorem~\ref{thm:pbseeger} \citep{seeger-02}, we have, with probability at least $1-\delta$ over $S\sim \Dcal^{n}$,
$$\Lcal_\Dcal (G_\theta) \leq p^*\,.$$
Then, we want to show
\begin{align} \label{eq:goal}
p^* = 
\inf_{C>0} \left\{\tfrac{1}{1-e^{-C}}\left(
    1-\exp\left(-C \, \widehat\Lcal_S (G_\theta) - \xi\right)
    \right)   \right\}.
\end{align}

\textbf{Case $\widehat\Lcal_S (G_\theta)< 1$:}
The function $\kl(\widehat\Lcal_S (G_\theta) \| p )$ is strictly increasing for $p>\widehat\Lcal_S (G_\theta)$. Thus, the supremum value $p^*$ is always reached in Equation~\eqref{eq:pstar}, so we have
\begin{align*} 
 \kl(\widehat\Lcal_S (G_\theta) \| p^* ) =  \xi,
\end{align*}
and, by Lemma~\ref{lem:supC=kl},
\begin{align}
    \sup_{C>0}   \Big[\Delta(C, \widehat\Lcal_S (G_\theta), p^*)\Big] 
    &= \xi\,, \label{aaa} \\
    \mbox{ and } \forall C > 0 : 
    \Delta(C, \widehat\Lcal_S (G_\theta), p^*)
    &\leq \xi\,.  \label{bbb}
\end{align}

Let 
$C^* \eqdef \argsup_{C>0}   \Big[\Delta(C, \widehat\Lcal_S (G_\theta), p^*)\Big]$. 

By rearranging the terms of $\Delta(C^*, \widehat\Lcal_S (G_\theta), p^*)$
(see Equation~\ref{eq:DeltaC}), we obtain, from Line~\eqref{aaa},
\begin{align} \label{aaa2}
    p^*  = \tfrac{1}{1-e^{-C^*}}\left(
     1-\exp\left(-C^* \, \widehat\Lcal_S (G_\theta) - \xi\right)\right),
\end{align}
and, from Line~\eqref{bbb},
\begin{align} \label{bbb2}
     \forall C > 0 :  p^*  \leq
     \tfrac{1}{1-e^{-C}}\left(
    1-\exp\left(-C \, \widehat\Lcal_S (G_\theta) - \xi\right)
    \right). 
\end{align}
Thus, combining Lines~\eqref{aaa2} and ~\eqref{bbb2}, we finally prove the desired result of Equation~\eqref{eq:goal}.

\textbf{Case $\widehat\Lcal_S (G_\theta)= 1$:} 
From Equation~\eqref{eq:pstar}, we have $p^* = 1$, because $\lim_{p\to 1} \kl(1\|p)=0$.
We also have $\frac{1-e^{-C-\xi}}{1-e^{-C}} \geq 1 $ and
$ \lim_{C\to\infty} \left[\frac{1-e^{-C-\xi}}{1-e^{-C}}\right]=1$, thus fulfilling Equation~\eqref{eq:goal}.
\end{proof}

\newpage
\section{Experiments}
\label{sec:supp_exp}

\subsection{Datasets}

In Section \ref{sec:xp} we use the datasets Ads (a small dataset related to advertisements on web pages), Adult (a low-dimensional task about predicting income from census data) and four binary variants of the MNIST handwritten digits:
\begin{description}
    \item[ads] http://archive.ics.uci.edu/ml/datasets/Internet+Advertisements\\
    The first 4 features which have missing values are removed.
    \item[adult] {https://archive.ics.uci.edu/ml/datasets/Adult}
    \item[mnist] http://yann.lecun.com/exdb/mnist/ \\
     Binary classification tasks are compiled with the following pairs: 
     \begin{itemize}
         \item Digits pairs 1 vs. 7, 4 vs. 9 and 5 vs. 6.
         \item Low digits vs. high digits $(\{0, 1, 2, 3, 4\}$ vs $\{5, 6, 7, 8, 9\})$ identified as \emph{mnistLH}.
     \end{itemize}
\end{description}

We split the datasets into training and testing sets with a 75/25 ratio. Table \ref{tab:datasets_overview} presents an overview of the datasets statistics.

\begin{table}[]
    \centering
    \caption{Datasets overview.}
    \label{tab:datasets_overview}
    \begin{tabular}{lrrr}
    \toprule
  
    Dataset & $n_{\rm train}$ &  $n_{\rm test}$ &  $d$\\
    \midrule
    ads     &   2459 & 820 & 1554 \\
    adult   &  36631 & 12211 & 108 \\
    mnist17 &   11377 & 3793 & 784\\
    mnist49 &   10336 & 3446 & 784 \\
    mnist56 &   9891 & 3298 & 784 \\
    mnistLH & 52500 & 17500 & 784 \\
    \bottomrule
\end{tabular}
\end{table}

\subsection{Learning algorithms details}
\label{sec:supp_algos}

Table~\ref{tab:models_overview} summarizes the characteristics of the learning algorithms used in the experiments.

\begin{table}[]
    \centering
    \caption{Models overview.}
    \label{tab:models_overview}
    \small
    \begin{tabular}{lccccc}
    \toprule
    Model name &   Cost function &  Train split &  Valid split & Model selection & Prior\\
    \midrule
    MLP   & linear loss, L2 regularized& 80\% & 20\% & valid linear loss & -\\
    PBGNet$_{\ell}$    & linear loss, L2 regularized & 80\% & 20\% & valid linear loss & random init\\
    PBGNet$_{\ell\text{-bnd}}$    & linear loss, L2 regularized & 80\% & 20\% & hybrid (see~\ref{sec:supp_algos})  & random init\\
    PBGNet   & PAC-Bayes bound & 100 \% & - & PAC-Bayes bound & random init\\
    \midrule
    PBGNet$_{\text{pre}}$ \\
    ~~-- pretrain & linear loss ($20$ epochs) & 50\% &- & - & random init\\
    ~~-- final & PAC-Bayes bound & 50\% & -  & PAC-Bayes bound & pretrain\\
    
    \bottomrule
\end{tabular}
\end{table}

\paragraph{Cost functions.}~\\ 
MLP, PBGNet$_{\ell}$, PBGNet$_{\ell\text{-bnd}}$\,: The gradient descent minimizes the following according to parameters~$\theta$.
\begin{equation*}
\widehat\Lcal_S (G_\theta) + \frac{\rho}{2}\|\theta\|^2\,,
\end{equation*}
where $\rho$ is the L2 regularization ``weight decay'' penalization term.

PBGNet, PBGNet$_{\text{pre}}$\,: The gradient descent minimizes the following according to parameters $\theta$ and $C\in\Rds^+$.
\begin{equation*}
    \frac{1}{1-e^{-C}}\left(
     1-\exp\left(-C \, \widehat\Lcal_S (G_\theta) - \frac1n \left[\Kcal (\theta, \mu)+\ln\tfrac{2\sqrt{n}}{\delta}\right]\right)
     \right),
\end{equation*}
where $\Kcal (\theta, \mu)$ is given by Equation~\eqref{eq:nonceKLnestpaspourlevieilhomme}.

\paragraph{Hyperparameter choices.} We execute each learning algorithm for combination of hyperparameters selected among the following values.
\begin{itemize}   
    \item Hidden layers $\in \{1, 2, 3\}$\,.
    \item Hidden size $\in \{10, 50, 100\}$\,. 
    \item Sample size $\in \{10, 50, 100, 1000, 10000\}$\,.
    \item Weight decay $\in \{0, 10^{-4}, 10^{-6}\}$\,.
    \item Learning rate $\in \{0.1, 0.01, 0.001\}$\,. 
\end{itemize}
 Note that the sample size does not apply to MLP and weight decay is set to 0 for PBGNet and PBGNet$_{pre}$).
 For the learning algorithms that use the PAC-Bayes bound for model selection, the union bound is applied to compute a valid bound value considering the 9 possible combinations of ``Hidden size'' and ``Hidden layers'' hyperparameters:
 we set $\delta = \frac{0.05}{9}$ in Theorem \ref{thm:bnd1to2} such that the selected model bound value holds with probability 0.95.

\begin{figure}
  \centering
  \includegraphics[trim={0cm 0cm 0cm 0cm},clip,width=\linewidth]{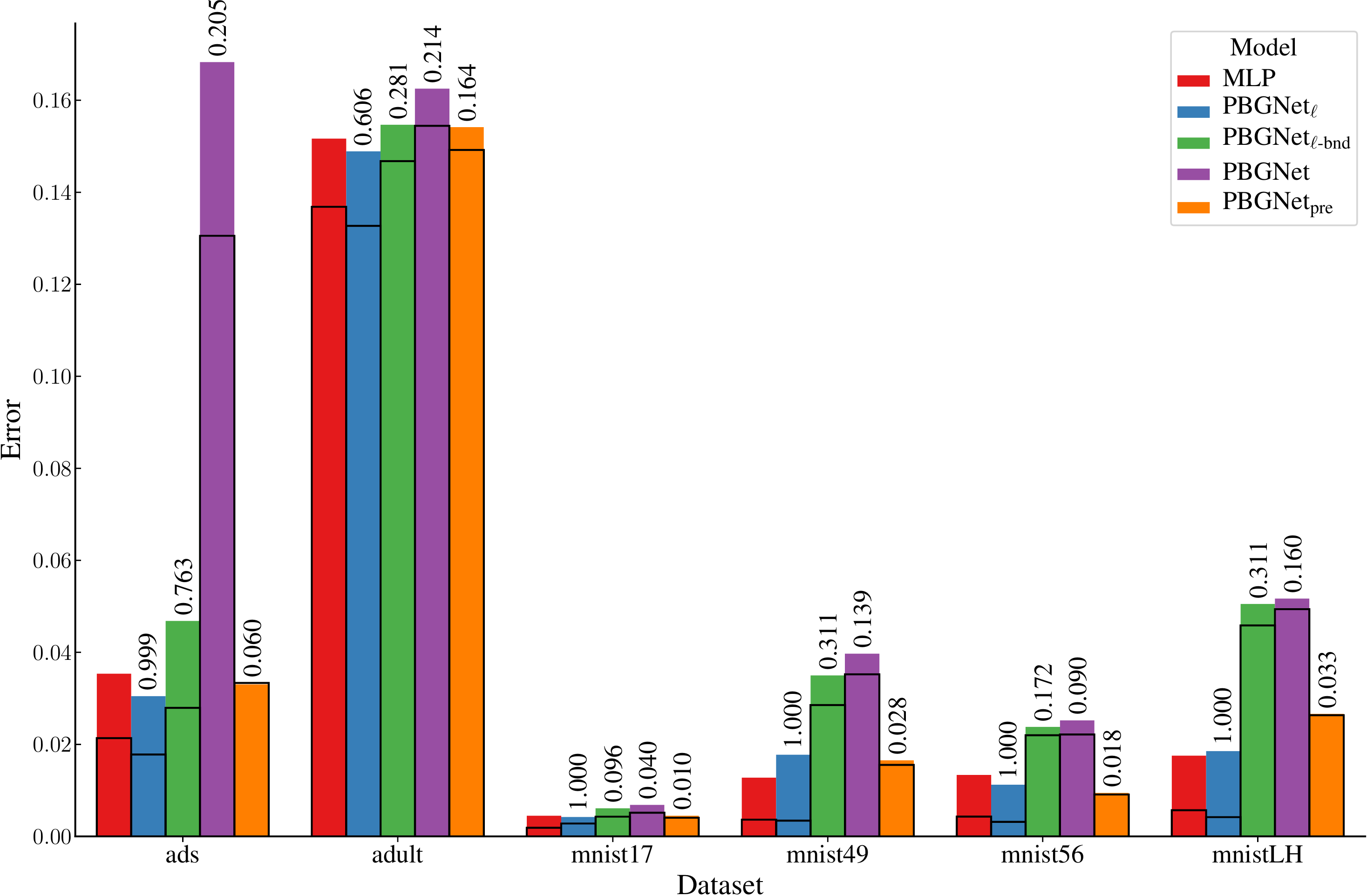}\llap{\makebox[5.5cm][l]{\raisebox{6.5cm}{\includegraphics[trim={3cm 19cm 11cm 1.5cm},clip,width=0.22
	\textwidth]{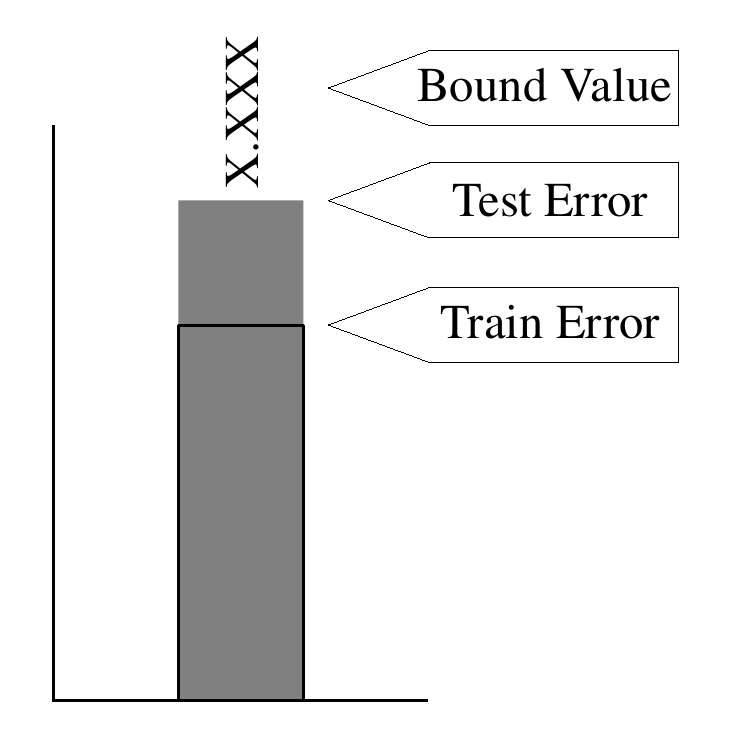}}}}
  \caption{Visualization of experiment results for the models on the binary classification datasets. 
  The colored bars display the test error while the black outlined bars exhibit the train error. 
  The PAC-Bayesian bounds are identified on the top of the bars and hold with probability 0.95.
  }
  \label{fig:res_sum_viz}
\end{figure}

 We report all the hyperparameters of selected models, for all learning algorithms and all datasets, in Table~\ref{tab:selected_models_overview}. The errors and bounds for these selected models are presented by Table~\ref{tab:main_results} of the main paper. The same results are visually illustrated by Figure~\ref{fig:res_sum_viz}.
 
\paragraph{Hybrid model selection scheme.}
Of note, PBGNet$_{\ell\text{-bnd}}$ has a unique hyperparameters selection approach using a combination of the validation loss and the bound value. 
First all hyperparameters, except the weight decay, are selected in order to minimize the bound value.
This includes choosing the best epoch from which loading the network weights.
Thus, we obtain the best models according to the bound for each weight decay values considered.
Then, the validation loss can be used to identify the best model between those, hence selecting the weight decay value. 


\paragraph{Optimization.}
For all methods, the network parameters are trained using the Adam optimizer \citep{kingma2014adam}
for a maximum of 150 epochs on mini-batches of size 32 for the smaller datasets (Ads and MNIST digit pairs) and size 64 for bigger datasets (Adult and mnistLH).
Initial learning rate is selected in $\{0.1, 0.01, 0.001\}$ and halved after each 5 consecutive epochs without a decrease in the cost function value.
We empirically observe that the prediction accuracy of PBGNet is usually better when trained using Adam optimizer than with \emph{plain} stochastic gradient descent, while both optimization algorithms give comparable results for our MLP model. The study of this phenomenon is considered as an interesting future research direction.

Usually in deep learning framework training loops, the empirical loss of an epoch is computed as the averaged loss of each mini-batch.
As the weights are updated after each mini-batch, the resulting epoch loss is only an approximation for the empirical loss of the final mini-batch weights.
The linear loss being a significant element of the PAC-Bayesian bound expression, the approximation has a non-negligible impact over the corresponding bound value.
One could obtain the accurate empirical loss for each epoch by assessing the network performance on the complete training data at the end of each epoch. We empirically evaluated that doing so leads to an increase of about a third of the computational cost per epoch for the inference computation.
A practical alternative used in our experiments is to simply rely on the averaged empirical loss on the mini-batches in the bound expression for epoch-related actions: learning rate reduction, early stopping and best epoch selection.

\paragraph{Prediction.}
Once the best epoch is selected, we can afford to compute the correct empirical loss for those weights and use it to obtain the corresponding bound value.
However, because PBGNet and its variants use a Monte Carlo approximation in the inference stage, the predicted output is not deterministic.
Thus, to obtain the values reported in Table~\ref{tab:main_results}, we repeated the prediction over the training data 20 times for the empirical loss computation of the selected epoch.
The inference repetition process was also carried out on the testing set, hence reported values $E_{S}$, $E_{T}$ and Bnd of the results consist in the mean over 20 approximated predictions.
The standard deviations are consistently below $0.001$, with the exception of PBGNet$_{\ell\text{-bnd}}$ on Ads for $E_{T}$ which has a standard deviation of 0.00165.
If network prediction consistency is crucial, one can set a higher sample size during inference to decrease variability, but keep a smaller sample size during training to reduce computational costs.

\pagebreak
\paragraph{Implementation details.} We implemented  PBGNet using \texttt{PyTorch} library \citep{paszke2017automatic}, using the \texttt{Poutyne} framework \citep{poutyne} for managing the networks training workflow. The code used to run the experiments is available at:
\begin{center}
   \url{https://github.com/gletarte/dichotomize-and-generalize}
\end{center}

When computing the full combinatorial sum, a straightforward implementation is feasible, gradients being computed efficiently by the automatic differentiation mechanism.
For speed purposes, we compute the sum as a matrix operation by loading all $\sbf\in\{-1,1\}^{d_{k}}$ as an array. 
Thus we are mainly limited by memory usage on the GPU, a single hidden layer of hidden size 20 using approximately 10Gb of memory depending on the dataset and batch size used.

For the Monte Carlo approximation, we need to insert the gradient approximation in a flexible way into the derivative graph of the automatic differentiation mechanism.
Therefore, we implemented each layer as a function of the weights and the output of the previous layer, with explicit forward and backward expression\footnote{See code in the following file:  \url{https://github.com/gletarte/dichotomize-and-generalize/blob/master/pbgdeep/networks.py}.}.
Thus the automatic differentiation mechanism is able to accurately propagate the gradient through our approximated layers, and also combine gradient from other sources towards the weights (for example the gradient from the KL computation when optimizing with the bound as the cost function).

Experiments were performed on NVIDIA GPUs (Tesla V100, Titan Xp, GeForce GTX 1080 Ti).

\begin{figure}
  \centering
  \includegraphics[trim={0.2cm 0.1cm 0.1cm 0cm},clip,width=\linewidth]{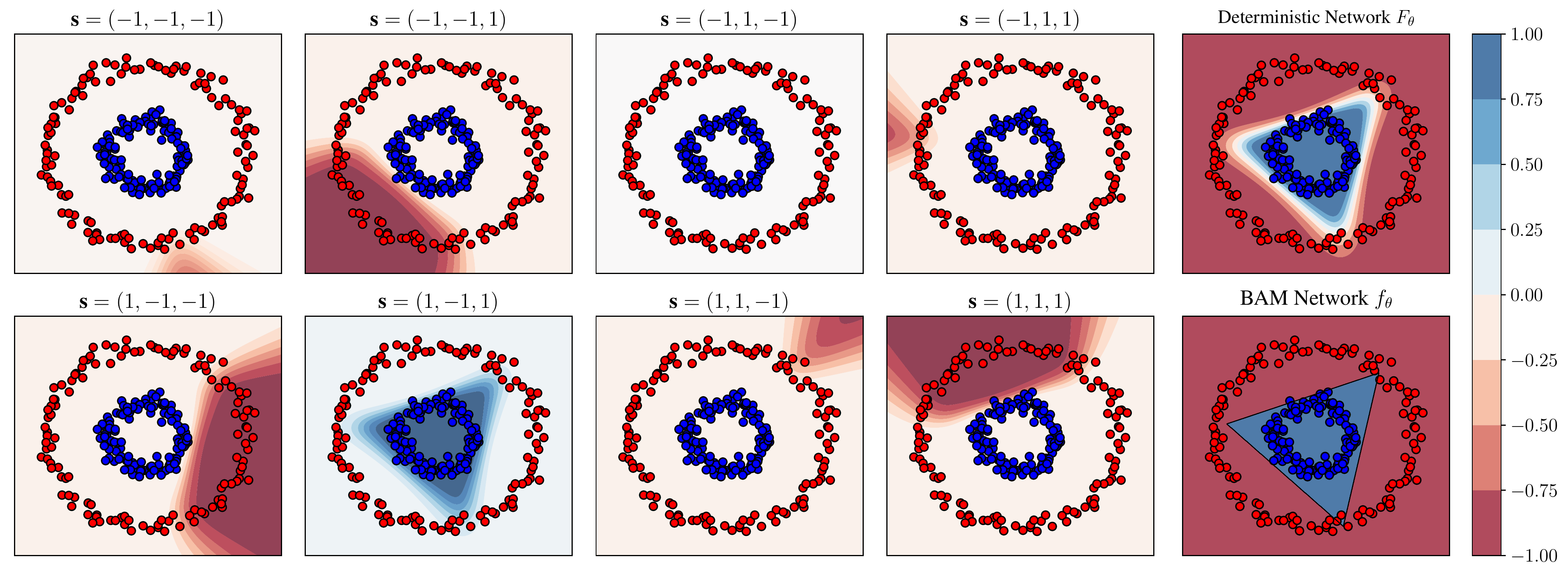}
  \caption{Illustration of the proposed method in Section \ref{sec:twolayers} for a one hidden layer network of size $d_1 = 3$, interpreted as a majority vote over $8$ binary representations $\sbf\in\{-1,1\}^{3}$. For each $\sbf$, a plot shows the values of $F_{\wbf_2}(\sbf) \Psi_\sbf(\xbf,\Wbf_1)$. The sum of these values gives the deterministic network output $F_\theta(\xbf)$ (see Equation~\ref{eq:F2layers}). We also show the BAM network output $f_\theta(\xbf)$ for the same parameters~$\theta$ (see Equation~\ref{eq:2BAM}).
  }
   \label{fig:circles}
\end{figure}

\begin{table}[]
    \centering
    \caption{Selected models overview.}
    \label{tab:selected_models_overview}
    \setlength{\tabcolsep}{3pt}
    {\normalsize
    \begin{tabular}{llccccccccc}
    \toprule
  
    Dataset & Model & Hid. layers &  Hid. size & $T$ & WD & C & KL & LR & Best epoch\\
    \midrule
    \multirow{5}{*}{ads} & MLP & 3 & 100 & - & 10$^{-4}$ & - & -  & 0.1 & 9\\
    & PBGNet$_{\ell}$ &  1 &  10 & 100 & 10$^{-6}$ &  11.44 & 14219 &  0.1 &   8 \\
    &  PBGNet$_{\ell\text{-bnd}}$ &  1 &   10 &  10 &  10$^{-6}$ &   4.47 &  2440 &  0.001 &   101 \\
    & PBGNet & 3 & 10 & 10000 & - & 0.47 & 27 &  0.01 & 49 \\
    & PBGNet$_{\text{pre}}$ &  3 & 10 & 10000 &  - & 0.58 & 0.09 &  0.1 & 82 \\
    \midrule
    \multirow{5}{*}{adult} & MLP & 2 & 100 & - & 0 & - & - & 0.1 & 21 \\
    & PBGNet$_{\ell}$ & 1 & 100 & 10000 & 10$^{-6}$ & 2.27 & 13813 & 0.1 & 25 \\
    & PBGNet$_{\ell\text{-bnd}}$ & 1 & 10 & 10 & 0 & 0.76 & 1294 & 0.001 	 & 111 \\
    & PBGNet & 1 & 10 & 1000 & - & 0.30 & 226 & 0.1 & 78 \\
    & PBGNet$_{\text{pre}}$ & 3 & 10 & 10000 & - & 0.09 & 0.13 & 0.01 & 73 \\
    \midrule
    \multirow{5}{*}{mnist17}& MLP & 2 & 50 & - & 0 & - & - & 0.01 & 56 \\
    & PBGNet$_{\ell}$ & 3 & 10 & 100 & 0 & 19.99 & 5068371 & 0.1 & 15 \\
    & PBGNet$_{\ell\text{-bnd}}$ & 1 & 10 & 10 & 10$^{-6}$ & 2.82 & 690 & 0.001 & 86 \\
    & PBGNet & 1 & 10 & 10000 & - & 1.33 & 164 & 0.1 & 106 \\
    &  PBGNet$_{\text{pre}}$ & 1 & 10 & 1000 & - & 0.73 & 0.46 & 0.1 & 71 \\
    \midrule
    \multirow{5}{*}{mnist49} & MLP & 2 & 100 & - & 10$^{-6}$ & - & - & 0.001 & 32 \\
    & PBGNet$_{\ell}$ & 2 & 50 & 10000 & 0 & 19.99 & 819585 & 0.01 & 33 \\
    & PBGNet$_{\ell\text{-bnd}}$ & 1 & 10 & 10 & 0 & 2.40 & 1960 & 0.001 & 102 \\
    & PBGNet & 1 & 10 & 10000 & - & 0.90 & 305 & 0.1 & 110 \\
    &  PBGNet$_{\text{pre}}$ & 1 & 100 & 10000 & - & 0.44 & 1.94 & 0.1 & 77 \\
    \midrule
    \multirow{5}{*}{mnist56} & MLP & 2 & 50 & - & 10$^{-6}$ & - & - & 0.001 & 17 \\
     & PBGNet$_{\ell}$ & 2 & 10 & 10000 & 0 & 19.99 & 883939 & 0.1 & 26 \\
     & PBGNet$_{\ell\text{-bnd}}$ & 1 & 10 & 10 & 0 & 1.95 & 808 & 0.001 & 55 \\
    & PBGNet & 1 & 10 & 10000 & - & 0.92 & 192 & 0.01 & 95 \\
    &  PBGNet$_{\text{pre}}$ & 1 & 50 & 10000 & - & 0.56 & 0.70 & 0.1 & 84 \\
    \midrule
    \multirow{5}{*}{mnistLH} & MLP & 3 & 100 & - & 10$^{-6}$ & - & - & 0.001 & 55 \\
    & PBGNet$_{\ell}$ & 3 & 100 & 10000 & 10$^{-6}$ & 19.99 & 98792960 & 0.1 & 92 \\
    & PBGNet$_{\ell\text{-bnd}}$ & 1 & 10 & 100 & 10$^{-6}$ & 2.00 & 8297 & 0.001 & 149 \\
    & PBGNet & 1 & 10 & 50 & - & 0.81 & 1544 & 0.1 & 107 \\
    & PBGNet$_{\text{pre}}$ & 2 & 100 & 10000 & - & 0.16 & 0.43 & 0.01 & 99 \\
    \bottomrule
    \end{tabular}
    }
\end{table}

\subsection{Additional results}

\begin{figure}[h]
  \centering
 \includegraphics[trim={1.2cm 1.1cm 0.5cm 0.4cm},clip,width=1\linewidth]{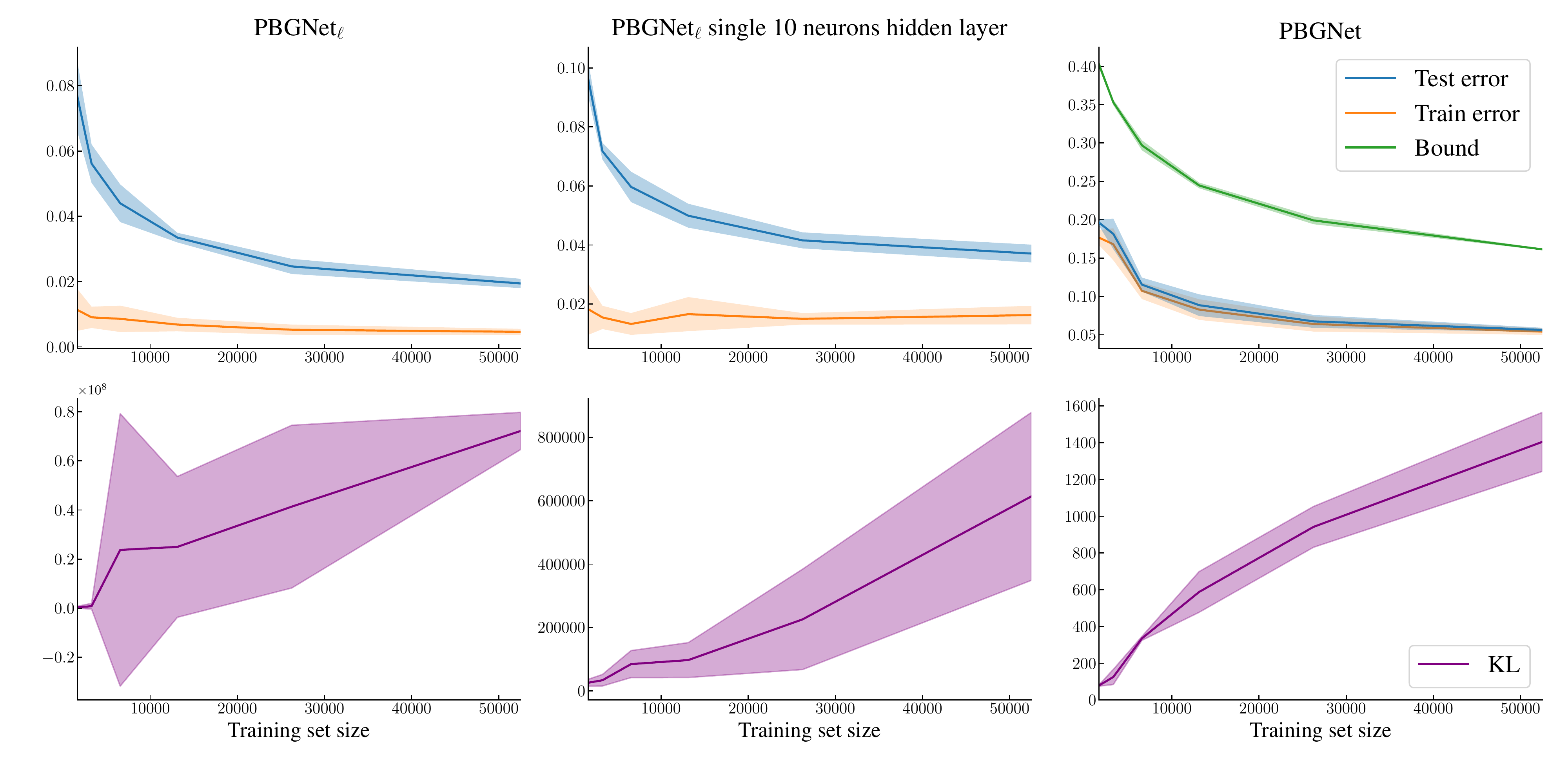}
  \caption{Study of the training sample size effect on PBGNet$_{\ell}$ and PBGNet for the biggest dataset, mnistLH. The middle column report results for PBGNet$_{\ell}$ with a fixed network architecture of a single hidden layer of 10 neurons, for a direct comparison with PBGNet which always selects this architecture.
  For each training set size values, 10 repetitions of the learning procedure with different random seeds were performed: each of them executed on different (random) train/test/valid dataset splits, and the stochastic gradient descent is initialized with different random weights. 
  Metrics means of the learned models are displayed by the bold line, with standard deviation shown with the shaded areas.
  Bound values for PBGNet$_{\ell}$ are trivial and thus not reported.
  We see that PBGNet bound minimization strikingly avoids overfitting by controlling the KL value according to the training set size. 
  On the opposite, PBGNet$_{\ell}$ achieves lower test risk, but clearly overfits small training sets.
  \vspace{-5mm}}  \label{fig:training_size} 
\end{figure}

Figure~\ref{fig:circles} reproduces the experiment presented by Figure~\ref{fig:moons} with another toy dataset. 
Table \ref{tab:error_bars} exhibits a variance analysis of Table \ref{tab:main_results}.
Figure~\ref{fig:training_size} shows the impact of the training set size.
Figure~\ref{fig:sample_effect} studies the effect of the sampling size $T$ on the stochastic gradient descent procedure. 
See figure/table captions for details.

\begin{table}[t]
  \caption{Variance analysis of the experiment presented in Table \ref{tab:main_results}. 
  We repeated 20 times the experimental procedure described in Section \ref{sec:xp}, but with a fixed network architecture of a single hidden layer of 10 neurons to limit computation complexity.
  Each repetition is executed on different (random) train/test/valid dataset splits, and the stochastic gradient descent is initialized with different random weights.
  The resulting standard deviations highlight the stability of the models.}
  \centering
  \setlength{\tabcolsep}{2.5pt}
  {\normalsize
  \begin{tabular}{l*{13}{c}}
    \toprule
    \multirow{2}{*}[-3pt]{Dataset} &  \multicolumn{2}{c}{MLP} & \multicolumn{2}{c}{PBGNet$_{\ell}$} & \multicolumn{3}{c}{PBGNet$_{\ell\text{-bnd}}$} & \multicolumn{3}{c}{PBGNet} & 
    \multicolumn{3}{c}{PBGNet$_{\text{pre}}$}\\
    \cmidrule(lr){2-3} \cmidrule(lr){4-5} \cmidrule(lr){6-8} \cmidrule(lr){9-11}\cmidrule(lr){12-14}
     & E$_{S}$ & E$_{T}$ & E$_{S}$ & E$_{T}$ & E$_{S}$ & E$_{T}$ & Bnd & E$_{S}$ & E$_{T}$ & Bnd & E$_{S}$ & E$_{T}$ & Bnd\\
    \midrule
    \multirow{3}{*}{ads} & 0.020  & 0.034 & 0.014  & 0.029 & 0.026 & 0.035 & 0.777 & 0.112 & 0.119 & 0.218 & 0.046 & 0.048 & 0.081\\\scriptsize
     & $\pm$ & $\pm$ & $\pm$ & $\pm$ & $\pm$ & $\pm$ & $\pm$ & $\pm$ & $\pm$ & $\pm$ & $\pm$ & $\pm$ & $\pm$\\
     &  0.005 & 0.007 & 0.003 & 0.008 & 0.003 & 0.005 & 0.000 & 0.003 & 0.007 & 0.002 & 0.007 & 0.012 & 0.007\\[0.25cm]
    \multirow{3}{*}{adult} & 0.133  & 0.149 & 0.132  & 0.148 & 0.148 & 0.151 & 0.271 & 0.156 & 0.159 & 0.215 & 0.153 & 0.154 & 0.166\\
     & $\pm$ & $\pm$ & $\pm$ & $\pm$ & $\pm$ & $\pm$ & $\pm$ & $\pm$ & $\pm$ & $\pm$ & $\pm$ & $\pm$ & $\pm$\\
     & 0.007 & 0.004 & 0.003 & 0.003 & 0.002 & 0.002 & 0.002 & 0.001 & 0.002 & 0.001 & 0.003 & 0.003 & 0.002\\[0.25cm]
    \multirow{3}{*}{mnist17} & 0.002  & 0.005 & 0.002  & 0.005 & 0.004 & 0.006 & 0.102 & 0.005 & 0.006 & 0.041 & 0.005 & 0.005 & 0.010\\
     & $\pm$ & $\pm$ & $\pm$ & $\pm$ & $\pm$ & $\pm$ & $\pm$ & $\pm$ & $\pm$ & $\pm$ & $\pm$ & $\pm$ & $\pm$\\
     & 0.001 & 0.001 & 0.000 & 0.001 & 0.000 & 0.001 & 0.004 & 0.000 & 0.001 & 0.001 & 0.001 & 0.001 & 0.001\\[0.25cm]
    \multirow{3}{*}{mnist49} & 0.003  & 0.013 & 0.004  & 0.016 & 0.031 & 0.033 & 0.300 & 0.039 & 0.040 & 0.143 & 0.019 & 0.019 & 0.031\\
     & $\pm$ & $\pm$ & $\pm$ & $\pm$ & $\pm$ & $\pm$ & $\pm$ & $\pm$ & $\pm$ & $\pm$ & $\pm$ & $\pm$ & $\pm$\\
     & 0.002 & 0.002 & 0.001 & 0.003 & 0.001 & 0.002 & 0.000 & 0.001 & 0.003 & 0.001 & 0.002 & 0.003 & 0.002\\[0.25cm]
    \multirow{3}{*}{mnist56} & 0.004  & 0.010 & 0.003  & 0.010 & 0.020 & 0.023 & 0.186 & 0.022 & 0.023 & 0.090 & 0.012 & 0.012 & 0.020\\
     & $\pm$ & $\pm$ & $\pm$ & $\pm$ & $\pm$ & $\pm$ & $\pm$ & $\pm$ & $\pm$ & $\pm$ & $\pm$ & $\pm$ & $\pm$\\
     & 0.001 & 0.002 & 0.001 & 0.002 & 0.001 & 0.002 & 0.000 & 0.001 & 0.002 & 0.001 & 0.002 & 0.003 & 0.002\\[0.25cm]
    \multirow{3}{*}{mnistLH} & 0.014  & 0.032 & 0.017  & 0.038 & 0.042 & 0.049 & 0.309 & 0.054 & 0.056 & 0.162 & 0.042 & 0.042 & 0.050\\
     & $\pm$ & $\pm$ & $\pm$ & $\pm$ & $\pm$ & $\pm$ & $\pm$ & $\pm$ & $\pm$ & $\pm$ & $\pm$ & $\pm$ & $\pm$\\
     & 0.003 & 0.002 & 0.001 & 0.003 & 0.001 & 0.003 & 0.001 & 0.001 & 0.002 & 0.001 & 0.002 & 0.002 & 0.002\\
    \bottomrule
\end{tabular}}
\label{tab:error_bars}
\end{table}

\afterpage{
\begin{figure}[h]
  \centering
 \includegraphics[trim={2cm 1cm 0.5cm 0.3cm},clip,width=1\linewidth]{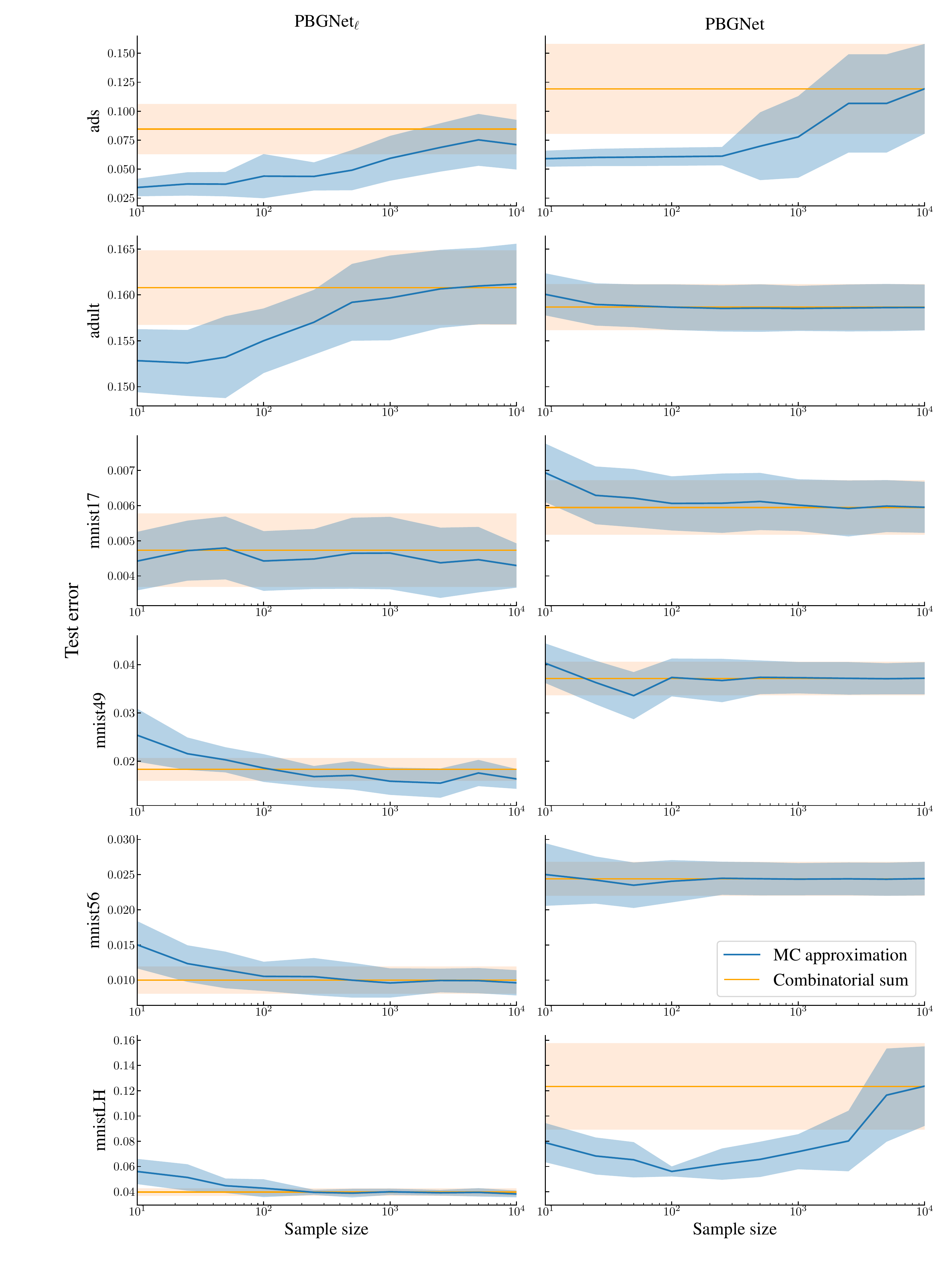}
  \caption{Impact of the sample size $T$ on stochastic gradient descent solution test error for PBGNet$_{\ell}$ and PBGNet. 
  Network parameters were fixed with a single hidden layer of size 10 and trained with initial learning rate of 0.1.
  For each sample size values and the combinatorial sum approach, 20 repetitions of the learning procedure with different random seeds were performed: each of them executed on different (random) train/test/valid dataset splits, and the stochastic gradient descent is initialized with different random weights. 
  The test error mean of the learned models is displayed by the bold line, with standard deviation shown with the shaded areas.\vspace{-5mm}}  \label{fig:sample_effect} 
\end{figure}
\clearpage
}

\end{document}